%% file: main.tex
\PassOptionsToPackage{hidelinks}{hyperref}
\documentclass{article}

\usepackage[preprint]{corl_2026} 

\input{preamble}

\title{
Grounding Generative Policies in Physics: Optimization-Guided Diffusion for Robot Control
}

\author{
  Sabrina Bodmer $^{*}$ \\
  ETH Zurich \\
  \And
  René Zurbrügg $^{*}$ \\
  ETH Zurich \\
  \And
  Tifanny Portela  \\
  ETH Zurich \\
  \AND
  Hao Ma   \\
  ETH Zurich \\
  \And
  Alexandre Didier\\
  ETH Zurich \\
  \And
  Marco Hutter\\
  ETH Zurich \\
  \AND
  Colin Jones $^{\dagger}$ \\
  EPFL \\
  \And
  Melanie Zeilinger $^{\dagger}$ \\
  ETH Zurich
\thanks{These authors contributed equally to this work. $\dagger$ Denotes equal supervision. \\ Corresponding authors: \texttt{sabodmer@ethz.ch, zrene@ethz.ch}}
}

\begin{document}
\addtocontents{toc}{\protect\setcounter{tocdepth}{-1}}
\maketitle


\begin{abstract}
Diffusion models sample effectively from high-dimensional, multimodal distributions, but their outputs may violate deployment constraints. For task-space robot policies, generated grasps, waypoints, or trajectories can be distributionally valid yet infeasible, violating reachability, collision-avoidance, or closed-loop executability requirements. This embodiment gap limits zero-shot deployment across robots, even when the task-space behavior itself is transferable.

We propose an inference-time optimization framework
that couples the behavior generation to physical feasibility by formulating diffusion guidance as a constrained optimization problem. Our key insight is to replace the sampling perturbation in the backward process with an optimized correction, allowing hard constraints or soft penalties to be imposed during sampling without the need to retrain the diffusion model, while keeping samples close to the learned prior.

We evaluate the method on dexterous grasp synthesis with reachability and collision-avoidance constraints, and dynamic manipulation with controller-level trackability constraints. Across settings and robot embodiments, optimization-guided denoising matches the feasibility of projection- and gradient-guidance baselines while better preserving grasp quality, and improving controller-level executability and task success, with task success improving by up to 20pp. on dexterous grasping and 23pp. on visuomotor manipulation over the best baseline.
\end{abstract}


\section{Introduction}
\label{sec:intro}
Diffusion models provide a flexible generative framework for robotics, capturing high-dimensional and multimodal distributions over actions, trajectories, or behaviors~\citep{chi2025diffusion,janner2022planning}.
Recent diffusion-based policies increasingly operate in task space rather than a specific robot's joint or motor space, predicting end-effector poses, waypoints, grasp candidates, or reference trajectories \cite{kim2024openvla, ha2024umi,punamiya2026egoverse}. This abstraction is attractive because it separates high-level behavior generation from robot-specific realization, including morphology, actuation, and low-level control. By providing a common interface, this approach allows policies to be transferred across entirely different embodiments without the costly need to collect new data or retrain models for each platform. However, task-space sampling only ensures training data consistency, not embodiment feasibility; predictions can violate kinematic reachability, joint/torque limits, collision-avoidance, or closed-loop execution constraints, transferring intent but failing on a specific physical robot.

Several lines of work have explored how diffusion-model outputs can be guided toward deployment constraints, including projection layers~\citep{christopher2024constrained}, differentiable barriers~\citep{ma25constraintawarediff}, constraint-aware training~\citep{li2024constraint}, gradient guidance~\citep{umionair2025}, and post-hoc projections~\citep{romer2025diffusion}. However, these approaches often require retraining, task-specific tuning, or setting-specific constraint formulations, and strong guidance can pull samples away from the learned prior. Cross-embodiment learning aims to share policies across robots through unified action spaces~\citep{openxembodiment2023,chi2024umi} or morphology-aware encoders~\citep{patel2024getzero}, but typically lacks an explicit inference-time mechanism for ensuring that sampled task-space predictions are executable on the target robot. This leaves open the need for a unified inference-time mechanism that keeps the generative prior fixed while applying minimal, structured corrections for embodiment, environment, and closed-loop execution constraints.

Our method is based on a structural property of DDIM sampling~\citep{DDIMsong2023denoising}: each reverse update separates the model-predicted denoising direction from a sampling perturbation. This separation provides a principled interface for constrained generation. Rather than modifying the pretrained score model or applying external gradients directly to the sample, we replace the perturbation with an optimization variable and regularize its magnitude, which keeps each step close to the pretrained sampling trajectory. The final denoising step can therefore satisfy user-defined constraints while preserving the behavior encoded by the pretrained prior. In this paper, we instantiate this principle for robot task-space policies, where the constraints encode reachability, collision avoidance, or controller-level executability. The resulting optimization-constrained denoising process injects information the diffusion model does not encode, without retraining or fine-tuning the model. As a result, the same task-space generative prior can be transferred zero-shot to different manipulator arms and environments by changing only the objectives of the optimization problem.

Concretely, our contributions are:
\begin{itemize}[leftmargin=1.5em]
    \item \textbf{Optimization-constrained denoising:} an inference-time framework that replaces the DDIM sampling perturbation with a constrained optimization variable, enabling small, structured corrections to a frozen diffusion prior.

    \item \textbf{Unified feasibility interface:} a formulation that can be solved either as a constrained nonlinear program or as a differentiable least-squares relaxation, with instantiations for reachability, collision avoidance, and controller-level executability.

    \item \textbf{Cross-embodiment generalization without retraining:} an evaluation showing that the same task-space diffusion prior can be reused across different simulated manipulator embodiments by changing only the inference-time optimization problem, improving grasp quality, task success, and controller-level executability.
\end{itemize}

\section{Related Work}
\label{sec:related}
\textbf{Cross-embodiment adaptation in diffusion policies.}
Cross-embodiment learning aims to reduce the need for large-scale, embodiment-specific demonstrations. One line of work learns from multi-robot datasets or shared action spaces: Open X-Embodiment aggregates demonstrations across robots to train RT-X models~\citep{openxembodiment2023}, Octo trains a transformer-based generalist policy with a diffusion action head on such aggregated data~\citep{team2024octo}, and UMI collects human demonstrations in an embodiment-agnostic end-effector space~\citep{chi2024umi}. Another line explicitly encodes morphology through hardware-conditioned, graph-based, modular, or transformer-based representations~\citep{chen2018hardware,wang2018nervenet,huang2020onepolicy,patel2024getzero}. These methods primarily address policy transfer, leaving embodiment-specific feasibility to be learned implicitly or handled by a downstream controller. UMI-on-Air is closest to our setting, guiding an embodiment-agnostic diffusion policy via low-level tracking costs~\citep{umionair2025}. In contrast, we treat the diffusion model as a fixed embodiment-agnostic prior and plug embodiment-specific feasibility into each reverse step through a replaceable constrained optimization module.

\textbf{Constraints in diffusion policies.}
Diffusion policies generally do not guarantee that generated predictions are executable for a specific embodiment or environment. Existing constraint-aware methods fall into two categories. The first targets environment-level safety or geometric feasibility, such as obstacle avoidance, collision avoidance, or path constraints. Projected diffusion models reformulate sampling as a constrained projection to enforce physical or geometric constraints~\citep{christopher2024constrained}, while other approaches penalize violations during training or employ barrier functions at inference~\citep{li2024constraint, ma25constraintawarediff}. Compositional methods such as Diffusion-CCSP and potential-based motion planning combine constraint energies to satisfy multiple task and environment constraints jointly~\citep{yang2023diffusionccsp,luo2024potential}. These methods improve task-level feasibility but typically do not capture embodiment dynamics, actuation limits, or controller response. The second category targets dynamic or control feasibility for a specific embodiment. DPCC combines model-based projection and constraint tightening during denoising to generate dynamically feasible control trajectories~\citep{romer2025diffusion}, while UMI-on-Air incorporates low-level controller tracking cost into sampling to favor trajectories that are easier to track~\citep{umionair2025}, and DynaGuide steers a frozen diffusion policy at inference by injecting the gradient of a learned latent dynamics model into each DDIM step, biasing samples toward outcomes that match user-provided goal images~\citep{du2026dynaguide}. These methods rely on soft guidance: feasibility is encouraged through denoising gradients, but corrections are not explicitly constrained, satisfaction can't be guaranteed and/or strong guidance can pull samples away from the learned prior. Multi-objective settings further rely on ad-hoc gradient aggregation, which can destabilize sampling when feasibility terms conflict. These limitations are shared by the broader family of gradient-based guidance, including classifier and classifier-free guidance, posterior sampling, and plug-and-play priors~\citep{graikos2022plugandplay,chung2023dps,bansal2023universal}, in which a cost gradient is added to each denoising step: feasibility is only encouraged rather than enforced, the guidance strength trades feasibility against fidelity, and constraint satisfaction cannot be guaranteed. In contrast, we replace the stochastic perturbation of each reverse step with a constrained optimization variable, so that feasibility is imposed as an explicit hard or soft constraint over the reverse process rather than as a soft gradient, while a regularizer on the correction keeps the sample close to the pretrained prior. This lets us enforce inverse-kinematic, joint-velocity, and controller-level feasibility.

\section{Background}
\label{sec:background}
Our method modifies the reverse sampling process of a pretrained diffusion model. We briefly review the DDIM sampler~\citep{DDIMsong2023denoising}, focusing on the stochastic perturbation that will later be replaced by an optimization variable.
A diffusion model~\citep{DDPMho2020denoising} corrupts a clean sample $x_0 \sim p_\text{data}$ as
\begin{align}\label{eq:forward}
    x_k = \sqrt{\alpha_k}\,x_0 + \sqrt{1-\alpha_k}\,\varepsilon,
    \qquad \varepsilon \sim \mathcal{N}(0,I),
\end{align}
and trains a denoising network $\hat{\varepsilon}_\theta(x_k,k)$ with the standard diffusion loss (Appendix~\ref{app:DDIM}). 
Given $x_k$, DDIM forms the clean-sample estimate $\hat{x}_0 = (x_k - \sqrt{1-\alpha_k}\,\hat{\varepsilon}_\theta(x_k,k))/\sqrt{\alpha_k}$ and performs the reverse update
\begin{align}\label{eq:ddim_reverse}
    x_{k-1}
    =
    \underbrace{
    \sqrt{\alpha_{k-1}}\,\hat{x}_0
    +
    \sqrt{1-\alpha_{k-1}-\sigma_k^2}\,
    \hat{\varepsilon}_\theta(x_k,k)
    }_{\textstyle \mu_\theta(x_k,k)}
    +
    \sigma_k\,\omega,
    \qquad
    \omega \sim \mathcal{N}(0,I),
\end{align}
with \(\sigma_k\) given in Appendix~\ref{app:DDIM}. This update separates the deterministic model prediction \(\mu_\theta(x_k,k)\) from the stochastic perturbation \(\omega\). We exploit this separation by replacing \(\omega\) with an optimization variable that enforces feasibility constraints during sampling.
\section{Method}
\label{sec:method}

\begin{figure*}[t!]
\vspace{-1cm}
    \includegraphics[width=1.\textwidth]{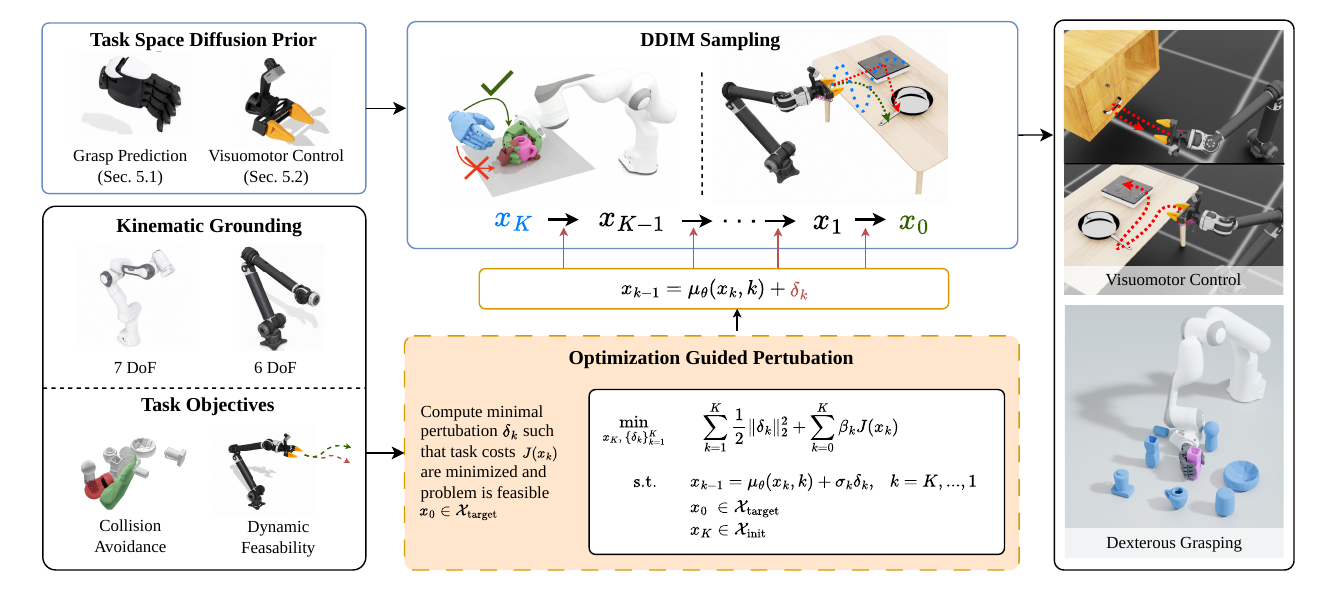}\caption{\textbf{Optimization-guided diffusion.}
A task-space diffusion prior generates an initial reverse denoising step. Instead of sampling the standard DDIM perturbation $\omega_k$, we replace it with a structured correction $\delta_k$ obtained from a constrained optimization problem. The objective minimizes the perturbation magnitude while incorporating embodiment- and environment-specific costs and constraints, such as $J$, $\mathcal{X}_{\text{target}}$, and $\mathcal{X}_{\text{init}}$. By changing only this constraint module, the same frozen prior can be reused across different manipulators.
}
    \label{fig:method_overview}
    \vspace{-5mm}
\end{figure*}

We present an inference-time framework to optimally guide a pretrained diffusion model. The key idea is to replace the stochastic perturbation in each DDIM reverse step with an optimized correction term. This preserves the nominal denoising direction predicted by the pretrained model, while using the correction term to enforce robot-, environment-, or task-specific feasibility. 
We use \(k=K,\dots,0\) to denote the diffusion denoising steps, where the subscripts $k=K$ and $k=0$ refers to the noisy sample and the clean prediction, respectively. Each diffusion variable \(x_k\) denotes an entire candidate output, such as a grasp or a reference trajectory, in which case $x_k = [x_{0|k}, \dots, x_{N|k}]$, where $n=0,\dots,N$ denotes the physical execution time along that trajectory.
\subsection{Optimization-Constrained Denoising}
\label{sec:method:opt}
The DDIM reverse update in~\eqref{eq:ddim_reverse} decomposes into a nominal prediction \(\mu_\theta(x_k,k)\) and a perturbation \(\sigma_k\omega\). We replace the random variable \(\omega\) with an optimization variable \(\delta_k\), such that
\begin{align}
    x_{k-1}
    =
    \mu_\theta(x_k,k) + \sigma_k \delta_k .
    \label{eq:guided_reverse}
\end{align}
The optimized perturbations \(\{\delta_k\}_{k=1}^{K}\) guide the reverse trajectory toward feasible outputs while regularizing deviations from the pretrained sampler.
Let \(J:\mathbb{R}^d \to \mathbb{R}_{\geq 0}\) be a feasibility cost, such as, e.g., a distance to the kinematically reachable set. We formulate optimization-guided sampling as
\begin{align}
\label{eq:opt}
\begin{aligned}
\min_{x_K,\,\{\delta_k\}_{k=1}^{K}} \quad
& \frac{1}{2}\sum_{k=1}^{K}  \|\delta_k\|_2^2
  + \sum_{k=0}^{K} \beta_{k} J(x_k) \\
\text{s.t.} \quad
& x_{k-1}
  =
  \mu_\theta(x_k,k) + \sigma_k \delta_k,
  && k=K,\dots,1, \\
& x_0 \in \mathcal{X}_{\mathrm{target}}, \\
& x_K \in \mathcal{X}_{\mathrm{init}}.
\end{aligned}
\end{align}
Problem~\eqref{eq:opt} admits a MAP interpretation in which \(\tfrac{1}{2}\|\delta_k\|_2^2\) is the negative log-density of a standard normal prior on \(\delta_k\) and \(\beta_k J(x_k)\) plays the role of a Boltzmann pseudo-likelihood on feasibility; we defer the derivation to Appendix~\ref{app:bayesian}. The terminal set \(\mathcal{X}_{\mathrm{target}}\) is typically chosen as a sublevel set of the feasibility cost, \(\mathcal{X}_{\mathrm{target}} = \{x \in \mathbb{R}^d : J(x) \leq \varepsilon_{\mathrm{tol}}\}\), enforcing the desired tolerance at the clean sample; \(\mathcal{X}_{\mathrm{init}}\) optionally restricts the initial noisy sample, with \(\mathcal{X}_{\mathrm{init}}=\mathbb{R}^d\) by default. The scaling by \(\sigma_k\) matches the correction scale to the stochastic term in the original DDIM sampler, so together with the Gaussian prior on \(\delta_k\) this encourages small, structured corrections from the learned reverse process. In practice, we evaluate feasibility costs only at denoising states close enough to the data manifold, since the cost gradients are often not meaningful for nearly Gaussian intermediate samples. This is done by increasing the penalty $\beta_k$ on the feasibility cost.

\subsection{Encoding Feasibility through \(J\)}
\label{sec:method:cost}
Problem~\eqref{eq:opt} is independent of the concrete feasibility model. Different robot-, environment-, or task-specific requirements can therefore be incorporated by changing \(J\) and the associated terminal set \(\mathcal{X}_{\mathrm{target}}\), while keeping the diffusion prior fixed. We consider three instantiations: kinematic reachability, collision avoidance, and controller-level executability. Platform-specific implementations are given in 
Appendix~\ref{app:implementation_details}.

\textbf{Kinematic Reachability.}
\label{sec:method:cost:ik}
Kinematic reachability captures whether a task-space output can be realized by the target embodiment within its admissible configuration space. Let \(\mathcal{X}_{\mathrm{IK}} \subseteq \mathbb{R}^d\) be the set of reachable task-space outputs. We define the ideal reachability cost as
\begin{align}
    J_{\mathrm{IK}}(x)
    =
    \min_{y \in \mathcal{X}_{\mathrm{IK}}}
    \|x-y\|_2 .
    \label{eq:Jik}
\end{align}
This cost vanishes on the reachable set and increases with distance from it. Using \(J=J_{\mathrm{IK}}\) in~\eqref{eq:opt} therefore guides sampling toward outputs that satisfy the kinematic limits of the target embodiment. The corresponding terminal set
\begin{align}
    \mathcal{X}_{\mathrm{target}}^{\mathrm{IK}}
    =
    \{\, x \in \mathbb{R}^d : J_{\mathrm{IK}}(x) \leq \varepsilon_{\mathrm{IK}} \,\}
    \label{eq:Xtarget:ik}
\end{align}
is a sublevel set of the cost $J_{\mathrm{IK}}$ and recovers exact reachability in the limit $\varepsilon_{\mathrm{IK}} \to 0$.
The framework does not assume a particular representation of $\mathcal{X}_{\mathrm{IK}}$. The cost may be computed using an analytical IK solver, a distance model, a projection operator,  or a learned differentiable surrogate. Alternatively, the cost could also be computed implicitly through a kinematic model constraint, i.e., using additional optimization variables.

\textbf{Collision Avoidance.}
\label{sec:method:cost:sdf}
A kinematically reachable output must additionally remain collision-free with respect to the environment and the robot itself. We encode collision avoidance through a signed distance function (SDF) $s : \mathbb{R}^m \to \mathbb{R}$, with $s(p) > 0$ on free space and $s(p) \leq 0$ inside obstacles. For a trajectory 
$x_k = [x_{0|k}, \dots, x_{N|k}]$, and safety margin $d_{\mathrm{safe}} \geq 0$, we define
\begin{align}
    J_{\mathrm{coll}}(x)
    =
    \sum_{n=0}^{N}
    \max\!\big(0,\; d_{\mathrm{safe}} - s(r_n)\big)^2.
    \label{eq:Jcoll}
\end{align}
The term \(J_{\mathrm{coll}}\) enforces the safety margin: it vanishes in free space beyond \(d_{\mathrm{safe}}\) and grows quadratically with the depth of any violation. 
Using \(J = J_{\mathrm{coll}}\) in~\eqref{eq:opt} therefore guides sampling toward collision-free outputs. 
The framework does not assume a particular representation of \(s\). It may be obtained from an analytical scene description, a precomputed voxel grid, a mesh-based query, or a learned neural SDF; gradient-based solvers additionally require \(s\) to be differentiable. Our implementation is described in Appendix~\ref{app:implementation_details}.

\textbf{Controller-Level Executability.}
\label{sec:method:cost:dyn}
Kinematic reachability and collision-avoidance alone do not ensure that a generated reference can be executed by the target robot under its low-level controller. We therefore also consider controller-level executability. Here, each diffusion variable represents a complete task-space reference trajectory, $x_k = [x_{0|k}, \dots, x_{N|k}]$,
where \(n=0,\dots,N\) indexes physical execution time.
Let \(\phi(x;q_0;\kappa)\) denote the closed-loop trajectory realizable when the robot starts from joint state \(q_0\) and tracks the reference \(x\) using a feasible task space controller $\kappa$. We define
\begin{align}
    J_{\mathrm{dyn}}(x)
    =
    \|x - \phi(x;q_0;\kappa)\|_2^2 .
    \label{eq:Jdyn}
\end{align}
This cost penalizes references whose commanded trajectory differs from what the closed-loop system can realize. Using \(J=J_{\mathrm{dyn}}\) in~\eqref{eq:opt} therefore guides sampling toward references that are executable by the target controller. 
The rollout map \(\phi\) depends on the low-level controller $\kappa$, reference-hold behavior, control frequency, and horizon. Our framework only requires that the cost can be evaluated or implicitly represented through additional optimization variables.  
The rollout model used in our dynamic manipulation experiments is described in Appendix~\ref{app:implementation_details}.

\textbf{Differentiable NLS (relaxation).}
\label{sec:method:nls}
When differentiability or runtime is critical, we relax~\eqref{eq:opt}
into a nonlinear least-squares problem over
\(z=\{x_K,\dots,x_0,\delta_K,\dots,\delta_1\}\). The reverse-step
equalities become squared consistency residuals, and the terminal
feasibility constraint is replaced by a hinge penalty
\(\beta_{\mathrm{term}}[J(x_0)-\varepsilon_{\mathrm{tol}}]_+^2\). 
We solve the NLS problem with
Theseus~\citep{pineda2022theseus} in a fixed small number of
Levenberg-Marquardt iterations. The relaxation enforces feasibility
only softly, but is differentiable and lightweight, which makes it
easier to embed in learning-based pipelines. The full objective and
weight choices are given in Appendix~\ref{app:nls}.

\section{Results}
\label{sec:results}
We evaluate our method on two settings. The first, dexterous grasping on two manipulator arms with a shared diffusion prior (Section~\ref{sec:result:grasp}), demonstrates gain over baselines, transfer across manipulator arms, and composition with collision constraints. The second, image-conditioned dynamic manipulation on two tasks (Section~\ref{sec:result:dyn}), tests extension from single-pose synthesis to controller-level trackability of full trajectories.

\textbf{Implementation Details.}
\label{sec:implementation}
The optimization interface in~\eqref{eq:opt} decouples the diffusion prior from
the particular feasibility model and numerical solver used at inference time.
We use three solver instantiations across our experiments: IPOPT for strict
terminal feasibility, Theseus for a differentiable nonlinear least-squares
relaxation, and L-BFGS for lightweight online replanning in the
image-conditioned manipulation setting. The corresponding feasibility costs,
closed-loop rollout approximations, solver settings, and relaxed objectives, as well as the hyperparameters for all baseline methods are
given in Appendix~\ref{app:implementation_details}.

\subsection{Dexterous grasping across manipulator arms}
\label{sec:result:grasp}
\begin{wraptable}{r}{0.6\textwidth}
\vspace{-4mm}
\input{tables/grasp}
\vspace{-4mm}
\end{wraptable}
We evaluate the framework on dexterous grasp synthesis with a 5-fingered XHand gripper. Each grasp is represented as a 21-dimensional configuration consisting of the wrist pose and 12 finger joints. A single task-space diffusion model from~\cite{zurbruegg2026dexevolve} is used unchanged across all experiments; embodiment-specific information enters only through the feasibility cost $J_{\mathrm{IK}}$. We deploy this prior on two manipulator arms with substantially different workspaces and joint limits, the Franka Panda~\citep{franka_panda} and the Dynaarm~\citep{duatic_dynaarm}. We predict $8$ grasps per object for $30$ objects across 5 randomized base poses spanning easy and hard reach configurations for a total of $1200$ grasps. \\
\textbf{Baselines.} \emph{DDIM$^\dagger$} reports the diffusion prior evaluated with no embodiment at all (floating gripper),
serving as an upper bound on what the generative model alone can produce when arm-reachability is removed from the problem. \emph{DDIM}~\citep{zurbruegg2026dexevolve} takes the raw diffusion output and solves for the closest IK-reachable grasp pose using cuRobo \footnote{\label{fn:srik_projection}For SR$^{\mathrm{IK}}$, we report the fraction of poses that are reachable before projection; after projection, this metric would be trivially 100\%.}~\citep{sundaralingam2023curobo}, representing a decoupled snap-to-feasibility baseline. \emph{Gradient Guidance}~\citep{umionair2025} applies classifier-style gradient updates of an IK cost along the reverse process; \emph{Projection Guidance}~\citep{romer2025diffusion} projects each denoising step onto the IK-feasible set during sampling. Our variants solve either the constrained problem~\eqref{eq:opt} (\emph{IPOPT}) or its differentiable relaxation~\eqref{eq:nls} (\emph{Theseus}). \\
\textbf{Metrics.}
$\mathrm{SR}^{(1)}$ reports per-(object, pose) success rates in simulation; $\mathrm{SR}^{\mathrm{IK}}$ is the fraction of kinematically reachable predictions; $E_{\mathrm{fc}}$, and $Q_1$ capture force-closure violation, and grasp wrench quality~\citep{graspqp2025}, respectively. Full definitions are given in Appendix~\ref{app:metrics}.

\subsubsection{Cross-embodiment performance} \vspace{-2mm}
\label{sec:result:grasp:main}


Table~\ref{tab:grasp_eval_updated} shows that without guidance, inverse-kinematic feasibility ($\mathrm{SR}^{\mathrm{IK}}$) is low on the tightly constrained Dynaarm workspace ($54.6\%$) and only moderate on the Panda ($83.7\%$). All guidance methods, including both baselines, raise $\mathrm{SR}^{\mathrm{IK}}$ above $95\%$. The differences appear in task success and grasp quality. On task success, projection guidance attains high feasibility but does not convert it into success ($\mathrm{SR}^{(1)}$ of $23.7\%$ and $37.4\%$ on the Dynaarm and Panda), whereas gradient guidance is the strongest baseline ($58.8\%$ and $50.9\%$). Our optimization-constrained approaches reach $63.5$--$69.8\%$ on the Dynaarm and $61.0$--$71.0\%$ on the Panda, exceeding gradient guidance by up to $11$ points on the Dynaarm and $20$ points on the Panda.
The pattern carries over to grasp quality: the cuRobo-snap DDIM baseline preserves feasibility by construction but collapses $Q_1$ ($\leq 7.1$ vs.\ $\geq14$ for our methods), and projection guidance similarly degrades $E_{\mathrm{fc}}$ and $Q_1$, consistent with the structural concern that projecting intermediate denoising states drags samples off the grasp manifold before the prediction becomes semantically meaningful.  The optimization-constrained approaches, by contrast, maintain $Q_1$ values approaching $\mathrm{DDIM}^\dagger$, showing that enforcing feasibility need not sacrifice grasp quality. The hard-constrained formulation using IPOPT achieves slightly higher success rates compared to the relaxed Theseus formulation, but at the cost of runtime: on the Dynaarm, DDIM denoising takes $79.3 \pm 1.3 \si{ms}$  per trajectory and gradient and projection guidance add negligible overhead ($84.2 \pm 0.9$ and $84.1\pm1.9\si{ms}$), whereas Theseus and IPOPT require $3.63\pm0.03 \si{s}$ and $5.47\pm0.6 \si{s}$, roughly $46\times$ and $69\times$ slower than DDIM.
We further provide evaluations broken down by base pose difficulty in Appendix~\ref{app:basepose},
 Figure~\ref{fig:easy_hard_poses}.
\subsubsection{Composing kinematic and collision constraints} \vspace{-2mm}
\begin{figure*}[t!]
    \centering
    \vspace{-5mm}
    \includegraphics[width=\linewidth]{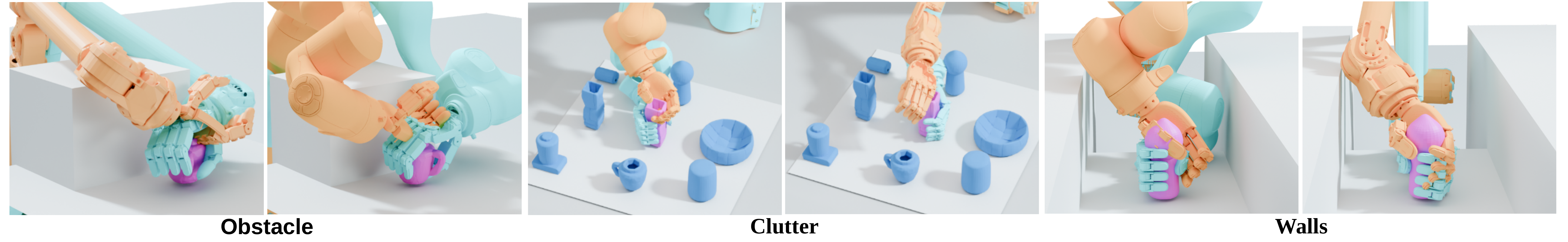}
\caption{\label{fig:collision_avoidance}
\textbf{Collision-Aware Grasping.} Grasp poses for different environments.
{\textcolor{graspBlue}{\rule{1.2ex}{1.2ex}}~IPOPT} respects collision constraints while preserving grasp quality; {\textcolor{graspOrange}{\rule{1.2ex}{1.2ex}}~Gradient Guidance} degrades the grasp.
}
    \vspace{-7mm}
     
\end{figure*}

\label{sec:result:grasp:collision}

We next add an environment-level collision cost $J_{\mathrm{coll}}$ to the feasibility module and evaluate our method in different constrained environments. The diffusion model and arm configurations remain unchanged; only the feasibility term is extended from $J_{\mathrm{IK}}$ to include $J_{\mathrm{coll}}$ in the cost and/or constraint formulation.
Figure~\ref{fig:collision_avoidance} illustrates qualitative failure modes: Gradient Guidance often produces collision-free grasps that fail to grasp the object, whereas our optimization-guided variants preserve stable grasps. Quantitative evaluations, including per-environment results, are provided in Appendix~\ref{app:collision}.

\subsection{Image-Conditioned Manipulation with a Floating-Gripper Prior}
\label{sec:result:dyn}
\begin{wrapfigure}{r}{0.65\textwidth}
\begin{minipage}{\linewidth}
\vspace{-5mm}
\centering
\captionof{table}{\textsc{Trajectory Predictions:} Comparison of methods across evaluation metrics, evaluated on 100 environments with statistics reported over 4 runs. DDIM$^\dagger$ denotes the raw diffusion model output deployed on a floating gripper without an arm. We compare ($x_k$)-nudging, which directly shifts the diffusion model output using the cost gradient, ($\sigma_k$)-guidance, which injects the gradient through the DDIM perturbation term ($\sigma_k \omega_k$), and our optimization-based guidance.}

\label{tab:tracking_eval_updated}
\resizebox{\linewidth}{!}{%
\setlength{\tabcolsep}{4pt}
\input{tables/vrdagger}
}
\end{minipage}
\vspace{-4mm}
\end{wrapfigure}
We next test whether optimization-guided denoising can make an
image-conditioned floating-gripper policy executable on embodied robot arms.
Unlike the grasping experiments, this setting requires steering short-horizon
trajectories that are repeatedly replanned from visual observations.\\
We train a Diffusion Transformer on demonstrations collected with a simulated,
modified Hoi!~\citep{engelbracht2025hoi} two-finger gripper in the Drawer Opening and Pick-and-Place
environments from~\citep{Zurbrugg2026VRDAggerIV} (Appendix~Fig.~\ref{fig:tasks}). The policy predicts task-space trajectories
of horizon length $N=6$ from image observations. After execution, a new
trajectory is generated based on the updated camera view.
At deployment, the policy is grounded to the Dynaarm or
Franka Panda through the controller-level executability cost
$J_{\mathrm{dyn}}$ from Section~\ref{sec:method:cost:dyn}. The cost compares
the predicted reference trajectory to a differentiable approximation of the
closed-loop trajectory under the low-level controller.  We use a Cartesian impedance controller as it provides passive compliance and admits an efficient differentiable closed-loop approximation. While MPC could improve tracking performance as in~\citep{umionair2025}, solving an MPC problem at each control step would dominate the inference budget. For online replanning, we optimize an additive trajectory offset during the final denoising steps using L-BFGS as a lightweight backend; implementation details are provided in Appendix~\ref{app:implementation_details}.
We compare against unguided DDIM and two gradient-guidance variants: an
$x_k$-nudging variant similar to~\citet{umionair2025}, and a
\emph{sigma-guidance} variant that replaces the sampling perturbation $\omega$
with a weighted gradient of the cost. Sigma guidance is the natural
single-gradient analogue of our method: it occupies the same reverse-step slot
that we instead fill with an optimized correction, so the gap between the two
isolates the benefit of optimizing the correction rather than merely injecting a
gradient. DDIM$^\dagger$ denotes the raw floating-gripper
policy without embodiment-level trackability constraints and serves as an
unembodied upper bound. We report task and subtask success rates on both
arms in Table~\ref{tab:tracking_eval_updated}.
On Drawer Opening, the methods are closely matched and the differences fall
within run-to-run variation: L-BFGS guidance moves task success from $79.5\%$
to $81.8\%$ on the Dynaarm and from $57.8\%$ to $60.0\%$ on the Franka,
comparable to the gradient-guidance baselines, with subtask success at
$87.3\%$ and $70.0\%$ respectively.
The gains are substantial on Pick and Place, where task success improves from
$26.0\%$ to $41.8\%$ on the Dynaarm and from $44.0\%$ to $67.0\%$ on the
Franka, while subtask success on the Franka approaches the unembodied upper
bound at $98.3\%$. On this harder task our method clearly outperforms unguided
DDIM and both gradient-guidance baselines on both embodiments, while sharing a
single floating-gripper prior.
These gains show that controller-level guidance corrects visually plausible
floating-gripper trajectories that are difficult for the embodied controller
to track.
The improvement comes at a moderate runtime cost: guided inference takes
$0.28$--$0.32$\,s compared to $0.02$--$0.07$\,s for unguided DDIM. Overall,
the results show that the same image-conditioned floating-gripper prior can be
deployed on embodied robot arms by changing only the inference-time
feasibility module, without retraining the policy.

\section{Limitations}
\label{sec:limitations}
Our framework has several limitations that suggest natural directions for future work. First, the constrained nonlinear program introduces non-trivial per-sample runtime, exceeding the gradient-guidance baselines; the Theseus and L-BFGS relaxations reduce this overhead but replace hard constraint satisfaction with soft penalties. Second, the problem in~\eqref{eq:opt} is generally nonconvex, so neither solver provides global feasibility or optimality guarantees, and we observed occasional convergence failures on poorly conditioned instances. Third, our main evaluation is conducted in simulation; we report preliminary hardware experiments in Appendix~\ref{app:real_world} as initial evidence that the same corrections transfer to physical deployment. Finally, our dynamic tracking cost assumes a Cartesian impedance controller and kinematic rollouts, an intentionally lightweight choice that shows the guidance mechanism helps even with a simple low-level controller. Replacing this approximation with a higher-fidelity model, such as an MPC or a differentiable physics simulator like Newton, is a promising direction but introduces additional design and numerical-conditioning challenges.

\section{Conclusion} 
\label{sec:conclusion}
We presented an inference-time framework to steer pretrained diffusion policies toward embodiment- and task-specific feasibility requirements. 
The key idea is to replace the reverse-step sampling perturbation with an optimization variable, allowing feasibility constraints to be imposed during denoising while leaving the generative model unchanged. 
By formulating diffusion guidance as a constrained optimization problem, our approach couples embodiment-agnostic behavior generation to physical feasibility at inference time, enabling a single diffusion prior to be reused across platforms by swapping only the optimization module.

We instantiated this interface for two complementary settings: kinematic reachability in dexterous grasp generation and controller-level trackability in dynamic manipulation. 
Across two manipulator arms, optimization-guided denoising achieved a better feasibility–success trade-off compared to projection- and gradient-guidance baselines, without retraining the diffusion model. 
These benefits come at a cost: the approach inherits the computational overhead of inference-time optimization, and the resulting nonconvex problems carry no global optimality guarantees.

Promising directions include reducing the solver runtime, extending the interface to richer contact and force constraints, and integrating optimization-guided denoising with large vision-language-action policies, where embodiment-specific retraining is particularly costly.


\newpage


\bibliography{main}  

\input{appendix_v2}

\end{document}

%% file: preamble.tex
\usepackage[T1]{fontenc}
\usepackage{times}

\usepackage{amsmath}
\usepackage{amssymb}
\usepackage{amsthm}
\usepackage{mathtools}
\usepackage{mathrsfs}
\usepackage{bbm}
\usepackage{siunitx}
\usepackage{booktabs}
\usepackage{multirow}
\usepackage{array}
\usepackage{makecell}
\usepackage{wrapfig}    

\usepackage{siunitx}
\usepackage{graphics}
\usepackage{graphicx}
\usepackage{adjustbox}
\usepackage{wrapfig}
\usepackage{svg}

\usepackage[table,dvipsnames]{xcolor}

\usepackage[font=small]{caption}
\usepackage{subcaption}
\usepackage{tikz}
\usetikzlibrary{fit,calc,positioning}

\usepackage{pgfplots}
\pgfplotsset{compat=1.16}

\usepackage{enumerate}
\usepackage{pbox}
\usepackage{dblfloatfix}
\usepackage{xspace}
\usepackage{multicol}
\usepackage{microtype}
\usepackage{bbding}

\usepackage[numbers]{natbib}

\usepackage[hidelinks]{hyperref}

\definecolor{mgreen}{RGB}{1,130,52}
\definecolor{mred}{RGB}{146,32,51}

\colorlet{mypink}{red!40}
\colorlet{myblue}{cyan!60}

\newcounter{oldtocdepth}

\newcommand{\noop}[1]{}
\newcolumntype{N}{S[table-format=3.1(1.1)]}
\usepackage{enumitem}
\definecolor{graspOrange}{HTML}{FFA83F}
\definecolor{graspBlue}{HTML}{52FAFF}
\usepackage{bm}
\usepackage{listings}
\usepackage[ruled,vlined]{algorithm2e}
\usepackage{tocloft}

%% file: tables/grasp.tex
\centering
\caption{\textsc{Grasp Predictions:} Comparison of methods across evaluation metrics, averaged over 30 objects, 8 grasps per object, and 5 base poses. DDIM$^\dagger$ is the raw diffusion model output with no IK checks or projection applied.}
\label{tab:grasp_eval_updated}
\resizebox{\linewidth}{!}{%
\setlength{\tabcolsep}{4pt}
\begin{tabular}{l|l||r|r|rr}
\toprule
\multirow{2}{*}{\textbf{Arm}} 
& \multirow{2}{*}{\textbf{Run}} 
& \multirow{2}{*}{\textbf{Success Rate}}
& \multirow{2}{*}{\textbf{Kinematics}} 
& \multicolumn{2}{c}{\textbf{Grasp}} \\
\cmidrule(lr){5-6}
& 
& $\text{SR}^{(1)}$ {\scriptsize$[\%]\ \uparrow$} 
& $\text{SR}^{\text{IK}}$ {\scriptsize$[\%]\ \uparrow$} 
& $E_{\text{fc}}$ {\scriptsize$(\times10^{3})\ \downarrow$} 
& $Q_1$ {\scriptsize$(\times10^{-3})\ \uparrow$} \\
\midrule 
& \cellcolor{blue!5}DDIM$^\dagger$
& \cellcolor{blue!5} $75.0$
& \cellcolor{blue!5} \ \ N/A
& \cellcolor{blue!5} $0.01$               & \cellcolor{blue!5} $16.0$ \\
\midrule \midrule
\multirow{5}{*}{\rotatebox{90}{Dynaarm}} & DDIM~\citep{zurbruegg2026dexevolve}
& $27.0$
& $54.6$
& $\ 2.37$                                 & $\ 6.5$ \\
& Projection Guidance
& $\ 23.7$
& $\mathbf{99.9}$
& $\ 0.72$                                 & $\ 6.8$ \\
& \cellcolor{gray!10}Gradient Guidance
& \cellcolor{gray!10} $58.8$
& \cellcolor{gray!10} \underline{$99.3$}
& \cellcolor{gray!10} $\ 0.97$             & \cellcolor{gray!10} $\ \ 12.80$ \\
\cmidrule(l){2-6} 
& Ours (Theseus)
& $\ \underline{63.5}$
& ${95.8}$
& $\ \underline{0.26}$                     & $\underline{14.5}$ \\
& \cellcolor{gray!10}Ours (IPOPT)
& \cellcolor{gray!10} $\mathbf{69.8}$
& \cellcolor{gray!10} ${96.4}$
& \cellcolor{gray!10} $\ \mathbf{0.20}$    & \cellcolor{gray!10} $\mathbf{15.2}$ \\
\midrule\midrule
\multirow{5}{*}{\rotatebox{90}{Franka Arm}}
& DDIM~\citep{zurbruegg2026dexevolve}
& $33.7$
& $83.7$
& $\ 2.02$                                 & $\ 7.1$ \\
& Projection Guidance
& $37.4$
& $\ \ \mathbf{100.0}$
& $\ 1.19$                                 & $\ 8.1$ \\
& \cellcolor{gray!10}Gradient Guidance
& \cellcolor{gray!10} $50.9$
& \cellcolor{gray!10} $97.9$
& \cellcolor{gray!10} $\ {1.04}$ & \cellcolor{gray!10} ${11.1}$ \\
\cmidrule(l){2-6}
& Ours (Theseus)
& $\underline{61.0}$
& $98.3$
& $\ \underline{0.23}$                                 & $\underline{14.1}$ \\
& \cellcolor{gray!10}Ours (IPOPT)
& \cellcolor{gray!10} $\mathbf{71.0}$
& \cellcolor{gray!10} $\underline{99.8}$
& \cellcolor{gray!10} $\ \mathbf{0.06}$    & \cellcolor{gray!10} $\mathbf{15.7}$ \\
\bottomrule
\end{tabular}
}

%% file: tables/vrdagger.tex
\begin{tabular}{c|l||rr||rr}
    \toprule
    \multirow{2}{*}{\textbf{Task}}
    & \multirow{2}{*}{\textbf{Method}}
    & \multicolumn{2}{c||}{\textbf{Dynaarm}}
    & \multicolumn{2}{c}{\textbf{Franka}} \\
    \cmidrule(lr){3-4} \cmidrule(lr){5-6}
    &
    & $\text{SR}_{\text{task}}$ {\scriptsize[\%]\,$\uparrow$}
    & $\text{SR}_{\text{sub}}$ {\scriptsize[\%]\,$\uparrow$}
    & $\text{SR}_{\text{task}}$ {\scriptsize[\%]\,$\uparrow$}
    & $\text{SR}_{\text{sub}}$ {\scriptsize[\%]\,$\uparrow$} \\
    \midrule
    \multirow{5}{*}{\rotatebox{90}{\textbf{Drawer}}}
     & \cellcolor{blue!5}DDIM$^\dagger$
       & \cellcolor{blue!5} 99.5{\scriptsize$\pm$1.0}
       & \cellcolor{blue!5} 100.0{\scriptsize$\pm$0.0}
       & \cellcolor{blue!5} {99.5{\scriptsize$\pm$1.0}}
       & \cellcolor{blue!5} {100.0{\scriptsize$\pm$0.0}} \\
    \cmidrule(l){2-6}
     & DDIM~\citep{zurbruegg2026dexevolve}
       & 79.5{\scriptsize$\pm$1.9}
       & 84.0{\scriptsize$\pm$0.8} 
       & 57.8{\scriptsize$\pm$2.2}
       & 66.3{\scriptsize$\pm$3.1} \\
     & \cellcolor{gray!10}Gradient Guidance ($\sigma$)
       & \cellcolor{gray!10} {79.8\scriptsize$\pm$1.7}
       & \cellcolor{gray!10} {85.0\scriptsize$\pm$1.6}
       & \cellcolor{gray!10} {57.8\scriptsize$\pm$7.8}
       & \cellcolor{gray!10} {66.8\scriptsize$\pm$6.8} \\
     & Gradient Guidance ($x_t$-nudge)
       & \underline{79.8}{\scriptsize$\pm$3.3}
       & \underline{83.8}{\scriptsize$\pm$3.0}
       & \underline{58.0}{\scriptsize$\pm$2.4}
       & \underline{67.0}{\scriptsize$\pm$2.8} \\
    \cmidrule(l){2-6} 
     & \cellcolor{gray!10}Ours (L-BFGS)
       & \cellcolor{gray!10} \textbf{81.8}{\scriptsize$\pm$2.9}
       & \cellcolor{gray!10} \textbf{87.3}{\scriptsize$\pm$2.9}
       & \cellcolor{gray!10} \textbf{60.0}{\scriptsize$\pm$3.4}
       & \cellcolor{gray!10} \textbf{70.0}{\scriptsize$\pm$1.6} \\
    \midrule\midrule
    \multirow{5}{*}{\rotatebox{90}{\textbf{Pick and Place}}}
     & \cellcolor{blue!5}DDIM$^\dagger$
       & \cellcolor{blue!5} {62.5\scriptsize$\pm$6.2}
       & \cellcolor{blue!5} {95.3\scriptsize$\pm$2.1}
       & \cellcolor{blue!5} {62.5\scriptsize$\pm$6.2}
       & \cellcolor{blue!5} {95.3\scriptsize$\pm$2.1}\\
    \cmidrule(l){2-6}
     & DDIM~\citep{zurbruegg2026dexevolve}
       & 26.0{\scriptsize$\pm$2.0}
       & 63.3{\scriptsize$\pm$3.3}
       & \underline{44.0}{\scriptsize$\pm$4.5}
       & 89.0{\scriptsize$\pm$3.4} \\
     & \cellcolor{gray!10}Gradient Guidance ($\sigma$)
       & \cellcolor{gray!10} \underline{29.3}{\scriptsize$\pm$4.3}
       & \cellcolor{gray!10} \textbf{69.3}{\scriptsize$\pm$4.6}
       & \cellcolor{gray!10} 41.0{\scriptsize$\pm$1.4}
       & \cellcolor{gray!10} \underline{92.3}{\scriptsize$\pm$1.3} \\
     & Gradient Guidance ($x_t$-nudge)
       & 26.3{\scriptsize$\pm$3.4}
       & \underline{63.3}{\scriptsize$\pm$5.0}
       & 43.3{\scriptsize$\pm$1.7} 
       & 90.3{\scriptsize$\pm$3.8} \\
    \cmidrule(l){2-6}
     & \cellcolor{gray!10}Ours (L-BFGS)
       & \cellcolor{gray!10} \textbf{41.8}{\scriptsize$\pm$3.8}
       & \cellcolor{gray!10} \underline{66.0}{\scriptsize$\pm$3.6}
       & \cellcolor{gray!10} \textbf{67.0}{\scriptsize$\pm$3.4}
       & \cellcolor{gray!10} \textbf{98.3}{\scriptsize$\pm$1.3} \\
    \bottomrule
    \end{tabular}

%% file: appendix_v2.tex
\newpage 
\addtocontents{toc}{\protect\setcounter{tocdepth}{3}}
\section{Supplementary Material}

\tableofcontents
\subsection{Denoising Diffusion Implicit Models}
\label{app:DDIM}
The following is based on the formulations of~\citep{DDIMsong2023denoising}. We consider a data distribution $q(x_0)$ and a model distribution $p_{\theta}(x_0)$, which approximates the data distribution and is easy to sample from. 
From $p_{\theta}(x)$ a sample $x_{k-1}$ can be generated from the previous sample $x_k$
\begin{align}
    x_{k-1} = \sqrt{\alpha_{k-1}}\left(\frac{x_k - \sqrt{1-\alpha_k}\hat\varepsilon_{\theta}^{(k)}(x_k)}{\sqrt{\alpha_k}}\right) + \sqrt{1- \alpha_{k-1}-\sigma_k^2} \hat\varepsilon_{\theta}^{(k)}(x_k) + \sigma_k\varepsilon_k, \quad \varepsilon_k \sim \mathcal{N}(0,I)
\end{align}

where $\alpha_k$ encodes how much original signal remains at timestep $k$ (i.e. is a noise-scheduler), and the parameters $\theta$ are then learned by minimizing the loss
\begin{align} \label{eq:ddim_loss}
    \mathcal{L}(\varepsilon_{\theta}) = \sum_{k=1}^K\mathbb{E}_{x_0 \sim q(x_0), \varepsilon_k \sim \mathcal{N}(0,I)} \left[||\varepsilon_{\theta}^{(k)}(\sqrt{\alpha_k}x_0 + \sqrt{1-\alpha_k}\varepsilon_k) - \varepsilon_k||_2^2\right], 
\end{align}
and
\begin{align} \label{eq:ddim_sigma}
    \sigma_k = \sqrt{\frac{1-\alpha_{k-1}}{1-\alpha_{k}}\left(1-\frac{\alpha_k}{\alpha_{k-1}}\right)}.
\end{align}

\subsubsection{Diffusion Models  Implementation Details}
We use existing diffusion-model architectures for both tasks. For grasp
synthesis, we use the Grasp Pose Diffusion model and training scheme from \citep{zurbruegg2026dexevolve}, omitting the keypoint embedding to reduce
inference time. For visuomotor servoing, we use the Diffusion Transformer with
a CNN image encoder from \citep{Zurbrugg2026VRDAggerIV}. The model is
conditioned on the two most recent RGB observations and hand poses, expressed
in the current gripper frame, and predicts a chunk of six future wrist poses and finger states in the same gripper frame. 

\newpage
\subsection{Bayesian interpretation of optimization-constrained denoising}
\label{app:bayesian}
This appendix shows that the cost terms in~\eqref{eq:opt} correspond to the negative log-density of a posterior over guided trajectories, justifying the L2 penalty on \(\delta_k\) as a Gaussian prior inherited from the pretrained reverse process.

\subsubsection{MAP Derivation}
\label{app:bayesian:map}
We view the guided denoising trajectory
\(\tau = (x_K, x_{K-1}, \dots, x_0)\), with each state \(x_k \in \mathbb{R}^d\), as a
latent variable and define a posterior over trajectories conditioned on feasibility:
\begin{align}
    p(\tau \mid \mathrm{feasible})
    \;\propto\;
    p_\theta(\tau)\prod_{k=1}^K \ell_k(x_k).
\end{align}
Here, \(p_\theta(\tau)\) denotes the trajectory distribution induced by the
pretrained DDIM sampler. The factors \(\ell_k(x_k)\) act as feasibility
pseudo-likelihoods and assign higher probability to lower-cost states:
\begin{align}
    \ell_k(x_k) \propto \exp\!\left(-\beta_k J(x_k)\right).
\end{align}
The cost \(J(x_k)\) measures violation of the desired feasibility criterion, and
\(\beta_k\) controls the strength of the guidance at denoising step \(k\). Thus,
states with smaller feasibility cost are exponentially preferred.

The pretrained prior factorizes along the reverse chain as
\begin{align}
    p_\theta(\tau) = p(x_K) \prod_{k=1}^K p_\theta(x_{k-1} \mid x_k).
\end{align}
Taking the negative logarithm yields
\begin{align}
\label{app:eq:neglogpost}
    -\log p(\tau \mid \mathrm{feasible}) = -\log p(x_K) - \sum_{k=1}^K \log p_\theta(x_{k-1} \mid x_k) + \sum_{k=1}^K \beta_{k} J(x_k) + c,
\end{align}
where, here and below, \(c\) collects all terms independent of the optimization
variables (such as the posterior normalizer) and may take a different value in
each equation. MAP inference corresponds to minimizing~\eqref{app:eq:neglogpost}
subject to the problem constraints: the terminal tolerance on \(x_0\) and the
initial-sample restriction on \(x_K\).

\subsubsection{Connection to DDIM}
\label{app:bayesian:ddim}
Under the stochastic DDIM update~\eqref{eq:guided_reverse} with \(\sigma_k > 0\),
the reverse kernel is Gaussian,
\begin{align}
    p_\theta(x_{k-1} \mid x_k) = \mathcal{N}\!\big(x_{k-1};\, \mu_\theta(x_k,k),\, \sigma_k^2 I\big),
\end{align}
so that
\begin{align}
    -\log p_\theta(x_{k-1} \mid x_k)
    = \frac{1}{2\sigma_k^2}\,\|x_{k-1} - \mu_\theta(x_k,k)\|_2^2
    + \tfrac{d}{2}\log(2\pi\sigma_k^2).
\end{align}
We reparameterize each transition through its injected noise,
\begin{align}
    x_{k-1} = \mu_\theta(x_k,k) + \sigma_k\,\delta_k,
    \qquad \delta_k \sim \mathcal{N}(0, I).
\end{align}
Substituting \(x_{k-1} - \mu_\theta(x_k,k) = \sigma_k\delta_k\) cancels the
\(\sigma_k^2\) and gives the identity
\begin{align}
    -\log p_\theta(x_{k-1} \mid x_k) = \tfrac{1}{2}\,\|\delta_k\|_2^2 + \tfrac{d}{2}\log(2\pi\sigma_k^2).
\end{align}
The second term is independent of \(\delta_k\). Summing over the chain and folding
all such terms into \(c\), \eqref{app:eq:neglogpost} becomes the MAP objective
\begin{align}
\label{app:eq:map_objective}
    -\log p(\tau \mid \mathrm{feasible})
    \;=\;
    -\log p(x_K)
    + \sum_{k=1}^K \tfrac{1}{2}\|\delta_k\|_2^2
    + \sum_{k=1}^K \beta_{k} J(x_k) + c.
\end{align}
This recovers the objective in~\eqref{eq:opt}, up to the prior term on \(x_K\) and the
replacement of the likelihood term on \(x_0\) by the terminal-tolerance constraint
used in our constrained formulation.

\subsection{Nonlinear Least-Squares Relaxation}
\label{app:nls}
We give the full objective used by the differentiable solver of
Section~\ref{sec:method:nls}. The optimization variables are the
denoising states and guidance perturbations,
\(z=\{x_K,\dots,x_0,\delta_K,\dots,\delta_1\}\), and we solve
\begin{align}
\label{eq:nls}
\min_{z}\quad
& \underbrace{\textstyle\sum_{k=1}^{K}\beta_\delta \|\delta_k\|_2^2}_{\text{guidance regularizer}}
+ \underbrace{\beta_{\mathrm{init}}\, d_{\mathcal{X}_{\mathrm{init}}}(x_K)^2}_{\text{initial-sample regularizer}}
\nonumber\\
&+ 
\underbrace{\textstyle\sum_{k=1}^{K}\beta_{\mathrm{rev}}\bigl\|x_{k-1}-\mu_\theta(x_k,k)-\sigma_k\delta_k\bigr\|_2^2}_{\text{reverse-process residuals}}
\nonumber\\
&+ \underbrace{\textstyle\sum_{k=1}^{K}\beta_{J,k} J(x_k)^2}_{\text{soft feasibility cost}}
+ \underbrace{\textstyle\sum_{k\in\mathcal{K}_J\setminus\{0\}}\beta_{\mathrm{path}}[J(x_k)-\varepsilon_{\mathrm{tol}}]_+^2
   + \beta_{\mathrm{term}}[J(x_0)-\varepsilon_{\mathrm{tol}}]_+^2}_{\text{soft feasibility constraints}},
\end{align}
with \([\cdot]_+=\max(0,\cdot)\). The initial-sample term
\(d_{\mathcal{X}_{\mathrm{init}}}(x_K)\) is replaced by a
regularizer \(\beta_{\mathrm{init}}\|x_K\|_2^2\) when no bounded
\(\mathcal{X}_{\mathrm{init}}\) is used. The reverse-process residuals
softly enforce consistency with the guided DDIM dynamics; the quadratic
\(\beta_{J,k} J(x_k)^2\) provides smooth gradients toward feasibility, while the hinge penalties concentrate weight on samples that exceed the tolerance. The path hinge is applied only to intermediate denoising states, and the separate terminal hinge \(\beta_{\mathrm{term}}\) on \(x_0\) reflects that final-output feasibility is the primary requirement. The last reverse step uses its own weight \(\beta_{\mathrm{rev},0}\) to couple \(x_0\) more tightly to the preceding denoising state.
The exact weights and full formulation are described in \ref{app:implementation_details}.

\addtocounter{figure}{1} 
\subsection{Visuomotor manipulation}
\label{app:visuomotor}

\begin{figure*}[t]
\centering
\includegraphics[width=1.0\textwidth]{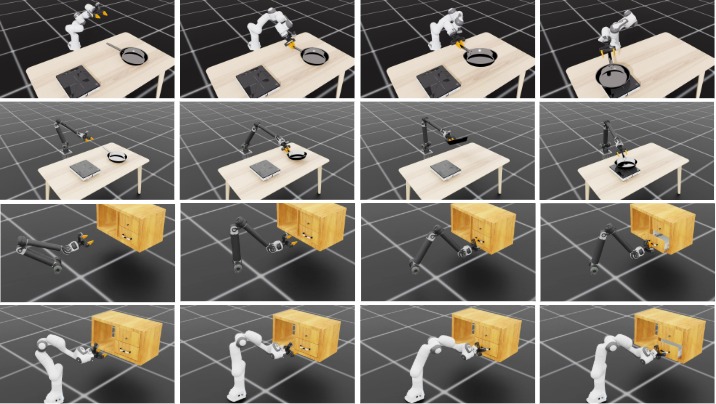}
\caption{\textbf{Example visuomotor manipulation tasks.}
We evaluate two image-conditioned manipulation tasks across two robotic manipulators.
The first two rows show tabletop pick-and-place, where the robot must grasp a pan and place it on the target burner.
The last two rows show drawer manipulation, where the robot must grasp the handle and open the drawer.}
\label{fig:tasks}
\end{figure*}

We evaluate visuomotor manipulation on two image-conditioned tasks,
shown in Fig.~\ref{fig:tasks}. In both tasks, the policy receives RGB
observations and predicts short-horizon end-effector reference trajectories,
which are executed on the target embodiment using a damped-least-squares
impedance controller. The differentiable rollout model used to evaluate
controller-level executability is described in Appendix~\ref{app:rollout};
implementation details for the corresponding guidance methods are provided in
Appendix~\ref{app:implementation_details}.

We report two success metrics. The task success rate
$\mathrm{SR}_{\mathrm{task}}$ measures the fraction of episodes in which the
full task is completed: the drawer is fully opened for drawer manipulation, and
the pan is correctly placed on the target burner for tabletop pick-and-place.
The subtask success rate $\mathrm{SR}_{\mathrm{sub}}$ gives partial credit for
meaningful intermediate progress. For drawer manipulation, the subtasks are
grasping the handle and opening the drawer by at least a small distance; for
pick-and-place, they are grasping the pan and lifting it from the table.

Table~\ref{tab:visuomotor_results} reports task success, subtask success,
executed end-effector distance, and solve time for both tasks and robot
embodiments. Across the embodiment-aware methods, the optimization-based
L-BFGS variant achieves the highest full task success in all settings. The
improvement is particularly pronounced for tabletop pick-and-place, where
successful execution often requires recovering from imperfect grasps or
regrasping the pan before placement. L-BFGS also generally solves successful
episodes faster than the baselines, despite the additional optimization step
during replanning.

\begin{table*}[t]
  \centering
  \caption{\textsc{Visuomotor manipulation results.} Success rates, executed end-effector distance, and completion time for drawer opening and tabletop pick-and-place on the Dynaarm and Franka. We report full task success $\mathrm{SR}_{\mathrm{task}}$, subtask success $\mathrm{SR}_{\mathrm{sub}}$, accumulated end-effector distance, and solve time. DDIM$^\dagger$ denotes the raw diffusion output deployed on a floating gripper without embodiment constraints and is shown as an upper-reference point rather than as an embodiment-aware method. Among embodiment-aware methods, our L-BFGS guidance achieves the highest task success in all settings and generally reduces solve time. Best values are \textbf{bold}, second-best are \underline{underlined}.}
  \label{tab:visuomotor_results}
  \resizebox{\textwidth}{!}{%
  \begin{tabular}{c|l||rrrr||rrrr}
    \toprule
    \multirow{2}{*}{\textbf{Task}}
    & \multirow{2}{*}{\textbf{Method}}
    & \multicolumn{4}{c||}{\textbf{Dynaarm}}
    & \multicolumn{4}{c}{\textbf{Franka}} \\
    \cmidrule(lr){3-6} \cmidrule(lr){7-10}
    &
    & $\text{SR}_{\text{task}}$ {\scriptsize[\%]\,$\uparrow$}
    & $\text{SR}_{\text{sub}}$ {\scriptsize[\%]\,$\uparrow$}
    & Dist {\scriptsize[m]\,$\downarrow$}
    & Time {\scriptsize[s]\,$\downarrow$}
    & $\text{SR}_{\text{task}}$ {\scriptsize[\%]\,$\uparrow$}
    & $\text{SR}_{\text{sub}}$ {\scriptsize[\%]\,$\uparrow$}
    & Dist {\scriptsize[m]\,$\downarrow$}
    & Time {\scriptsize[s]\,$\downarrow$} \\
    \midrule
    \multirow{5}{*}{\rotatebox{90}{\textbf{Drawer}}}
     & \cellcolor{blue!5}DDIM$^\dagger$
       & \cellcolor{blue!5} 99.5{\scriptsize$\pm$1.0}
       & \cellcolor{blue!5} 100.0{\scriptsize$\pm$0.0}
       & \cellcolor{blue!5} 0.66{\scriptsize$\pm$0.00}
       & \cellcolor{blue!5} 7.67{\scriptsize$\pm$0.10}
       & \cellcolor{blue!5} 99.5{\scriptsize$\pm$1.0}
       & \cellcolor{blue!5} 100.0{\scriptsize$\pm$0.0}
       & \cellcolor{blue!5} 0.66{\scriptsize$\pm$0.00}
       & \cellcolor{blue!5} 7.67{\scriptsize$\pm$0.10} \\
    \cmidrule(l){2-10}
     & DDIM~\citep{zurbruegg2026dexevolve}
       &  79.5{\scriptsize$\pm$1.9}
       &  84.0{\scriptsize$\pm$0.8}
       &  \underline{0.68}{\scriptsize$\pm$0.01}
       &  \underline{9.00}{\scriptsize$\pm$0.06}
       &  57.8{\scriptsize$\pm$2.2}
       &  66.3{\scriptsize$\pm$3.1}
       &  \textbf{0.69}{\scriptsize$\pm$0.00}
       &  \underline{10.06}{\scriptsize$\pm$0.43} \\
     & \cellcolor{gray!10}Gradient Guidance ($\sigma$)
       & \cellcolor{gray!10} \underline{79.8}{\scriptsize$\pm$1.7}
       & \cellcolor{gray!10} \underline{85.0}{\scriptsize$\pm$1.6}
       & \cellcolor{gray!10} 0.69{\scriptsize$\pm$0.01}
       & \cellcolor{gray!10} 9.03{\scriptsize$\pm$0.18}
       & \cellcolor{gray!10} 57.8{\scriptsize$\pm$7.8}
       & \cellcolor{gray!10} 66.8{\scriptsize$\pm$6.8}
       & \cellcolor{gray!10} 0.70{\scriptsize$\pm$0.00}
       & \cellcolor{gray!10} 10.29{\scriptsize$\pm$0.42} \\
     & Gradient Guidance ($x_t$-nudge)
       &  79.8{\scriptsize$\pm$3.3}
       &  83.8{\scriptsize$\pm$3.0}
       &  \textbf{0.68}{\scriptsize$\pm$0.01}
       &  9.09{\scriptsize$\pm$0.04}
       &  \underline{58.0}{\scriptsize$\pm$2.4}
       &  \underline{67.0}{\scriptsize$\pm$2.8}
       &  \underline{0.70}{\scriptsize$\pm$0.01}
       &  10.21{\scriptsize$\pm$0.43} \\
    \cmidrule(l){2-10}
     & \cellcolor{gray!10}Ours (L-BFGS)
       & \cellcolor{gray!10} \textbf{81.8}{\scriptsize$\pm$2.9}
       & \cellcolor{gray!10} \textbf{87.3}{\scriptsize$\pm$2.9}
       & \cellcolor{gray!10} 0.73{\scriptsize$\pm$0.01}
       & \cellcolor{gray!10} \textbf{8.12}{\scriptsize$\pm$0.19}
       & \cellcolor{gray!10} \textbf{60.0}{\scriptsize$\pm$3.4}
       & \cellcolor{gray!10} \textbf{70.0}{\scriptsize$\pm$1.6}
       & \cellcolor{gray!10} 0.78{\scriptsize$\pm$0.02}
       & \cellcolor{gray!10} \textbf{9.42}{\scriptsize$\pm$0.44} \\
    \midrule\midrule
    \multirow{5}{*}{\rotatebox{90}{\textbf{Pick and Place}}}
     & \cellcolor{blue!5}DDIM$^\dagger$
       & \cellcolor{blue!5} 62.5{\scriptsize$\pm$6.2}
       & \cellcolor{blue!5} 95.3{\scriptsize$\pm$2.1}
       & \cellcolor{blue!5} 1.70{\scriptsize$\pm$0.05}
       & \cellcolor{blue!5} 7.39{\scriptsize$\pm$0.07}
       & \cellcolor{blue!5} 62.5{\scriptsize$\pm$6.2}
       & \cellcolor{blue!5} 95.3{\scriptsize$\pm$2.1}
       & \cellcolor{blue!5} 1.70{\scriptsize$\pm$0.05}
       & \cellcolor{blue!5} 7.39{\scriptsize$\pm$0.07} \\
    \cmidrule(l){2-10}
     & DDIM~\citep{zurbruegg2026dexevolve}
       &  26.0{\scriptsize$\pm$2.0}
       &  63.2{\scriptsize$\pm$3.3}
       &  1.21{\scriptsize$\pm$0.01}
       &  \underline{9.33}{\scriptsize$\pm$0.20}
       &  \underline{44.0}{\scriptsize$\pm$4.5}
       &  89.0{\scriptsize$\pm$3.4}
       &  \underline{0.98}{\scriptsize$\pm$0.04}
       &  \underline{8.94}{\scriptsize$\pm$0.35} \\
     & \cellcolor{gray!10}Gradient Guidance ($\sigma$)
       & \cellcolor{gray!10} \underline{29.3}{\scriptsize$\pm$4.3}
       & \cellcolor{gray!10} \textbf{69.3}{\scriptsize$\pm$4.6}
       & \cellcolor{gray!10} \underline{1.21}{\scriptsize$\pm$0.01}
       & \cellcolor{gray!10} 10.08{\scriptsize$\pm$0.73}
       & \cellcolor{gray!10} 41.0{\scriptsize$\pm$1.4}
       & \cellcolor{gray!10} \underline{92.3}{\scriptsize$\pm$1.3}
       & \cellcolor{gray!10} 1.00{\scriptsize$\pm$0.00}
       & \cellcolor{gray!10} 8.99{\scriptsize$\pm$0.32} \\
     & Gradient Guidance ($x_t$-nudge)
       &  26.3{\scriptsize$\pm$3.4}
       &  63.2{\scriptsize$\pm$5.0}
       &  \textbf{1.19}{\scriptsize$\pm$0.01}
       &  9.56{\scriptsize$\pm$0.57}
       &  43.3{\scriptsize$\pm$1.7}
       &  90.3{\scriptsize$\pm$3.8}
       &  \textbf{0.97}{\scriptsize$\pm$0.01}
       &  9.22{\scriptsize$\pm$0.33} \\
    \cmidrule(l){2-10}
     & \cellcolor{gray!10}Ours (L-BFGS)
       & \cellcolor{gray!10} \textbf{41.8}{\scriptsize$\pm$3.8}
       & \cellcolor{gray!10} \underline{66.0}{\scriptsize$\pm$3.6}
       & \cellcolor{gray!10} 1.46{\scriptsize$\pm$0.03}
       & \cellcolor{gray!10} \textbf{8.40}{\scriptsize$\pm$0.49}
       & \cellcolor{gray!10} \textbf{67.0}{\scriptsize$\pm$3.4}
       & \cellcolor{gray!10} \textbf{98.3}{\scriptsize$\pm$1.3}
       & \cellcolor{gray!10} 1.21{\scriptsize$\pm$0.01}
       & \cellcolor{gray!10} \textbf{8.02}{\scriptsize$\pm$0.41} \\
    \bottomrule
  \end{tabular}

  }
\end{table*}
The executed-distance metric should be interpreted together with task success. Methods that solve more episodes also complete more of the reaching, recovery, and placement motions, which can naturally increase accumulated path length. Thus, the longer paths observed for L-BFGS reflect its ability to recover from near-failure cases, including through corrective motions such as regrasping, while still completing successful episodes faster overall.

\subsection{Optimization Modules} \label{app:optimization_modules} \setcounter{subsubsection}{-1} The optimization-guided samplers require differentiable modules that translate robot-specific execution constraints into costs that can be optimized during denoising. We use two such modules. For grasp synthesis, a learned kinematic surrogate approximates the distance of a task-space wrist target to the reachable set. For visuomotor manipulation, a differentiable closed-loop rollout model approximates how the low-level controller tracks a predicted reference trajectory. This section describes both modules, including their inputs, training or rollout construction, and how their outputs are used as guidance costs. \subsubsection{Kinematic Feasibility Surrogate} \label{app:ik_surrogate} For the grasping experiments in Section~\ref{sec:result:grasp}, we approximate the distance-to-reachability cost \(J_{\mathrm{IK}}\) with a differentiable surrogate \(\hat{J}_{\mathrm{IK}}\). The surrogate replaces repeated calls to an IK solver during denoising with a cheap neural approximation, while still providing gradients with respect to the generated grasp pose. In addition to a scalar reachability estimate, the network predicts a corrective task-space offset and a joint configuration used for forward-kinematics consistency during training. 

  \paragraph{Network architecture.}
  The surrogate is a feed-forward MLP that maps a single task-space
  wrist target to a reachability estimate and a corrective offset. The target is
  parameterized as a $9$-D vector $\bm{p}=[\,\bm{t}\in\mathbb{R}^3,\;\bm{r}_{6\mathrm{D}}\in\mathbb{R}^6\,]$,
  where $\bm{t}$ is the wrist position in the robot base frame and $\bm{r}_{6\mathrm{D}}$ is the continuous
  $6$-D rotation representation (the first two columns of the rotation matrix, with the
  full matrix recovered by Gram--Schmidt).
  \begin{algorithm}[t]
  \caption{Kinematic feasibility surrogate $\hat{J}_{\mathrm{IK}}$: architecture and training}
  \label{alg:ik_mlp}
  \DontPrintSemicolon
  \SetKwInOut{Const}{Const}
  \SetKwProg{Fn}{Function}{:}{end}
  \SetKwFunction{Net}{IKMLPNet}
  \SetKwFunction{IKSolver}{IKSolver}
  \SetKwFunction{KP}{KeypointMSE}
  \Const{$H=256$,\ dof $\in\{6,7\}$,\ act $=$ GELU;\ epochs $=5$,\ batch $=4096$,\ lr $=3\!\times\!10^{-4}$;\ 
  $\lambda_{\mathrm{err}}=10,\ \lambda_{\mathrm{pose}}=1,\ \lambda_{\mathrm{fk}}=1$}
  \Fn{\Net{$\bm{p}$}}{
    \tcp{$\bm{p}=[\text{pos}(3),\ \text{rot6d}(6)]$; quat input converted}
    $\bm{h} \leftarrow \mathrm{GELU}(\mathrm{Linear}(9,256)(\bm{p}))$ \tcp*{input embedding}
    $\bm{h} \leftarrow \mathrm{GELU}(\mathrm{Linear}(256,256)(\bm{h}))$ \tcp*{backbone block 1}
    $\bm{h} \leftarrow \mathrm{GELU}(\mathrm{Linear}(256,256)(\bm{h}))$ \tcp*{backbone block 2}
    $\hat{\bm{q}} \leftarrow \mathrm{Linear}(128,\text{dof})(\mathrm{GELU}(\mathrm{Linear}(256,128)(\bm{h})))$ \tcp*{joint config (for
  FK loss)}
    $\hat{J}_{\mathrm{IK}} \leftarrow \mathrm{softplus}(\mathrm{Linear}(64,1)(\mathrm{GELU}(\mathrm{Linear}(256,64)(\bm{h}))))$
  \tcp*{reachability $\ge 0$}
    $\Delta\bm{p} \leftarrow \mathrm{Linear}(128,9)(\mathrm{GELU}(\mathrm{Linear}(256,128)(\bm{h})))$ \tcp*{pose correction}
    \KwRet $\hat{\bm{q}},\ \hat{J}_{\mathrm{IK}},\ \Delta\bm{p}$\;
  }
  \tcp{Data: $10^5$ targets labeled offline by reference IK solver}
  sample $\bm{p} = [\mathrm{pos}, \mathrm{rot}]\sim \mathcal{U}([-2,2]^3) \times SO(3)$

$(\bm{q}^\star,\,J^\star,\,\bm{p}^{\mathrm{reach}}) \leftarrow$ \IKSolver{$\bm{p}$}\;  \tcp{Pre-Compute Dataset}

$\theta \leftarrow$ init \Net;\ opt $\leftarrow$ AdamW;\ sched $\leftarrow$ CosineAnnealing\; 
  \For{epoch $=1$ \KwTo $50$}{
    \For{minibatch $(\bm{p},\,J^\star,\,\bm{p}^{\mathrm{reach}})$}{
      $\hat{\bm{q}},\ \hat{J}_{\mathrm{IK}},\ \Delta\bm{p} \leftarrow$ \Net{$\bm{p}$}\;
      
      $\mathcal{L}_{\mathrm{err}} \leftarrow \lVert \hat{J}_{\mathrm{IK}} - J^\star \rVert^2$\;
      
      $\mathcal{L}_{\mathrm{pose}} \leftarrow$ \KP{$\bm{p}+\Delta\bm{p}$, $\bm{p}^{\mathrm{reach}}$}\;
      
      $\mathcal{L}_{\mathrm{fk}} \leftarrow$ \KP{$\mathrm{FK}(\hat{\bm{q}})$, $\bm{p}^{\mathrm{reach}}$}\;

      $\mathcal{L} \leftarrow \lambda_{\mathrm{err}}\mathcal{L}_{\mathrm{err}} + \lambda_{\mathrm{pose}}\mathcal{L}_{\mathrm{pose}} +
  \lambda_{\mathrm{fk}}\mathcal{L}_{\mathrm{fk}}$\;
  
      opt.step$(\nabla_\theta \mathcal{L})$\;
    }
    sched.step()\;
  }
  \KwRet $\theta$\;
  \end{algorithm}

  A quaternion input ($\bm{t}+$ \texttt{wxyz})
  is accepted and converted internally.
  The backbone is an input embedding $\mathrm{Linear}(9\!\to\!256)$ followed by two
  hidden blocks of width $256$; every block is $\mathrm{Linear}\!\to\!\mathrm{GELU}$ with
  no normalization and no dropout. Three heads branch from the final $256$-D feature:
  \begin{itemize}[leftmargin=1.5em,itemsep=2pt]
    \item \textbf{Reachability head} $\hat{J}_{\mathrm{IK}}$:
  $\mathrm{Linear}(256\!\to\!64)\!\to\!\mathrm{GELU}\!\to\!\mathrm{Linear}(64\!\to\!1)$
    followed by a \texttt{softplus}, which guarantees a \emph{non-negative scalar distance}
    $\hat{J}_{\mathrm{IK}}(\bm{p})\ge 0$ rather than a feasibility logit. It regresses the
    residual pose error of the reference IK solver; reachability is decided by the
    threshold $\hat{J}_{\mathrm{IK}}\le\varepsilon_{\mathrm{IK}}$.
    \item \textbf{Pose-correction head} $\Delta\bm{p}\in\mathbb{R}^9$: an \emph{additive}
    offset (in the same $[\bm{t},\bm{r}_{6\mathrm{D}}]$ space) that nudges an infeasible
    target toward the reachable set; it supplies the guidance gradient during denoising.
    \item \textbf{Configuration head} $\hat{\bm{q}}\in\mathbb{R}^{\mathrm{dof}}$: predicts a
    joint solution, used for the forward-kinematics consistency loss and for visualization.
  \end{itemize}
  The network has $\approx 2.2\times10^{5}$ parameters ($\approx\!218$k for the $6$-DoF arm,
  $\approx\!219$k for the $7$-DoF arm). Linear layers use Kaiming-normal initialization.

\paragraph{Training-data generation.}
We generate training data separately for each embodiment, namely the Franka and
Dynaarm. For each sample, we draw the target position uniformly from a
$[-2,2]^3$\,m box around the robot base and sample the target orientation from
a normalized Gaussian quaternion, $\bm{q}\sim\mathcal{N}(0,I_4)$. Each target is
labeled by solving an inverse-kinematics (IK) problem, which returns the closest
reachable pose and the corresponding joint configuration. For each embodiment,
we generate $10^5$ training targets and $2\times10^3$ validation targets. 

\begin{table}[t]
    \centering
    \caption{\textbf{Kinematic feasibility surrogate accuracy.}
Held-out accuracy of $\hat{J}_{\mathrm{IK}}$ on uniformly sampled task-space targets.
RMSE reports the error relative to the ground-truth reachability cost $J_{\mathrm{GT}}$.
Ranking AUC reports how well $-\hat{J}_{\mathrm{IK}}$ predicts cuRobo IK solvability under a $5\,\mathrm{mm}/0.10\,\mathrm{rad}$ tolerance.
Both arms achieve millimeter-level prediction error, and high AUC values show that lower surrogate costs reliably rank targets as more reachable.}
    \label{tab:ik_surrogate}
    \begin{tabular}{lcc}
      \toprule
      Embodiment & $\hat{J}_{\mathrm{IK}}$ RMSE [mm] & Ranking AUC \\
      \midrule
      Franka ($7$ DoF)   & $2.9$ & $0.993$ \\
      Dynaarm ($6$ DoF) & $5.2$ & $0.930$ \\
      \bottomrule
    \end{tabular}
  \end{table}

  \paragraph{Training objective and optimization.}
  Training minimizes a weighted \emph{regression} objective (no classification loss):
  \begin{equation}
    \mathcal{L} \;=\;
    \lambda_{\mathrm{err}}\,\big\|\hat{J}_{\mathrm{IK}}-J^{\text{GT}}\big\|_2^2
    \;+\;\lambda_{\mathrm{pose}}\,\mathcal{L}_{\text{pose}}
    \;+\;\lambda_{\mathrm{fk}}\,\mathcal{L}_{\text{fk}}
    \;+\;\lambda_{\bm{q}}\,\big\|\hat{\bm{q}}-\bm{q}^\star\big\|_2^2 ,
  \end{equation}
  where $\mathcal{L}_{\text{pose}}$ is a keypoint MSE between the corrected pose
  $\bm{p}+\Delta\bm{p}$ and the reachable reference pose, and $\mathcal{L}_{\text{fk}}$ is a
  forward-kinematics consistency MSE between $\mathrm{FK}(\hat{\bm{q}})$ and that reference
  pose. We use $\lambda_{\mathrm{err}}=10$, $\lambda_{\mathrm{pose}}=1$,
  $\lambda_{\mathrm{fk}}=1$, and $\lambda_{\bm{q}}=0$ (direct joint supervision is off; the
  configuration head is still trained implicitly through $\mathcal{L}_{\text{fk}}$). Since
  $\lambda_{\bm{q}}=0$, the joint-supervision term vanishes and the loss reduces to the
  three-term objective used in Algorithm~\ref{alg:ik_mlp}.
  Optimization uses AdamW with learning rate $3\times10^{-4}$, a cosine-annealing schedule,
  batch size $4096$, $2-5$ epochs, and fixed seed $42$.
The resulting held-out surrogate accuracy is reported in
Table~\ref{tab:ik_surrogate}.

\subsubsection{Closed-Loop Rollout Model}
\label{app:rollout}

This section specifies the differentiable closed-loop rollout map
\(\phi(x;q_0;\kappa)\) used to evaluate the controller-level executability cost
\(J_{\mathrm{dyn}}\)~\eqref{eq:Jdyn} in the dynamic-manipulation experiments
(Section~\ref{sec:result:dyn}). The map predicts how a planned end-effector
trajectory would actually be executed by the controller, so that
\(J_{\mathrm{dyn}}\) can score the executed poses against the planned references.
Its inputs are
\begin{itemize}
  \item the \(N\) predicted end-effector waypoints
        \(x=[x_{0|k},\dots,x_{N-1|k}]\), where each
        \(x_{t|k}=(p^{\mathrm{ref}}_t,R^{\mathrm{ref}}_t)\in\mathrm{SE}(3)\) is
        expressed in the frame \(E_0\) of the end-effector at replanning time \(k\);
  \item the measured arm configuration \(q_0\in\mathbb{R}^{n}\) (with \(n=6\) for the
        Dynaarm and \(n=7\) for the Franka); and
  \item the fixed Cartesian impedance controller parameters \(\kappa\),
\end{itemize}
and its output is the sequence of planned \emph{executed} poses
\(\hat x_t=\phi(x;q_0;\kappa)_t\).
\paragraph{Per-step controller update.}
We approximate the closed-loop controller with a kinematic rollout. The model assumes that gravity compensation is handled by feed-forward torques, and simulates only the tracking behavior of a damped-least-squares inverse-kinematics controller. At each reference step $t$, the rollout starts from the current joint configuration $q_t$, computes the end-effector pose
$T^W_E(q_t)=(p^{ee}_t,R^{ee}_t)$, and updates the joints through the following four steps.

\begin{enumerate}
\item \textbf{Task-space error.}
We first compute the geometric pose error between the reference pose and the current end-effector pose:
\begin{equation*}
\Delta x_t
=
\begin{bmatrix}
p^{\mathrm{ref}}_t - p^{ee}_t \\
\omega^{\mathrm{err}}_t
\end{bmatrix}
\in\mathbb{R}^6,
\qquad
\omega^{\mathrm{err}}_t
=
\log\big(R^{\mathrm{ref}}_t R^{ee,\top}_t\big)^{\vee}
\in\mathbb{R}^3 .
\end{equation*}
Here, $\log(\cdot)^{\vee}:\mathrm{SO}(3)\to\mathbb{R}^3$ denotes the logarithmic map from rotations to axis-angle vectors. Optionally, this error can be weighted by a diagonal task-space stiffness $K_p^{\mathrm{task}}$. In all reported experiments, we set $K_p^{\mathrm{task}}=I$, so $\Delta x_t$ is the unweighted geometric pose error.

\item \textbf{Resolved-rate joint update.}
The task-space error is mapped to a joint-space increment with a damped-least-squares resolved-rate update:
\begin{equation*}
\delta q_t
=
J(q_t)^{\top}
\big(
J(q_t)J(q_t)^{\top}+\lambda^2 I_6
\big)^{-1}
\Delta x_t,
\qquad
\lambda=0.05 ,
\end{equation*}
where $J(q_t)$ is the geometric end-effector Jacobian.

\item \textbf{Authority limits.}
Before applying the update, we clip the joint increment to the motion that the robot can realize within one reference step. The per-joint bound is
\begin{equation*}
\bar{\delta q}
=
\min
\left(
\dot q_{\max}\,\Delta t_{\mathrm{ref}},
\frac{\tau_{\max}}{k_p^{\mathrm{joint}}}
\right),
\end{equation*}
where $\dot q_{\max}$ and $\tau_{\max}$ are the robot's velocity and effort limits, and $k_p^{\mathrm{joint}}$ is the joint-space PD stiffness specified by the robot model. This bound captures both velocity limits and finite torque authority.

\item \textbf{PD lag and integration.}
Finally, we account for the fact that the low-level PD controller closes only part of the commanded joint-space gap during one reference step. We model this with a first-order lag factor $\alpha_{\mathrm{eff}}$ and integrate the clipped increment:
\begin{equation*}
q_{t+1}
=
\mathrm{clip}
\Big(
q_t
+
\alpha_{\mathrm{eff}}\odot
\mathrm{clip}(\delta q_t,\pm\bar{\delta q}),
\underline q,
\overline q
\Big),
\end{equation*}
with
\begin{equation*}
\alpha_{\mathrm{eff},j}
=
1
-
\left(
1
-
\frac{k_{p,j}^{\mathrm{joint}}}
{k_{p,j}^{\mathrm{joint}} + k_{d,j}^{\mathrm{joint}}/\Delta t_{\mathrm{in}}}
\right)^{n_{\mathrm{sub}}}.
\end{equation*}
Here, $k_{d}^{\mathrm{joint}}$ denotes the joint-space damping gains from the robot model, $\Delta t_{\mathrm{in}}$ is the inner control timestep, and $n_{\mathrm{sub}}$ is the number of inner controller steps per reference interval. The final clipping enforces the joint limits $\underline q$ and $\overline q$.
\end{enumerate}

\paragraph{Executability cost.}
Given the executed poses \(\hat x_t=\phi(x;q_0;\kappa)_t\), the executability
cost~\eqref{eq:Jdyn} is the step-weighted tracking residual
\begin{equation}
\label{eq:appeqdin}
  J_{\mathrm{dyn}} \;=\;
   \!\sum_{t=0}^{N-1} c_t\left(\,\big\|\,\hat p_t - p^{\mathrm{ref}}_t\big\|^2 + 
  \big\|\,(\hat R_t - R^{\mathrm{ref}}_t)\,\mathcal{K}\,\big\|^2 \right),
\end{equation}
where the orientation term is scored through three orthogonal keypoints
\(\mathcal{K}\) at distance \(0.15\,\mathrm{m}\). The position and orientation
terms are equally weighted, and the step weights increase linearly as
\(c_t=t{+}1\).

\paragraph{Controller offset.}
Using~\eqref{eq:appeqdin} directly as a guidance signal can produce an undesirable feedback effect.
Since $J_{\mathrm{dyn}}$ penalizes tracking error, the easiest way to reduce this cost is not necessarily to make the motion more executable, but to make the reference itself less demanding. This occurs because the actions passed to the impedance controller also define the target reference used to evaluate tracking. As a result, the gradient can reduce the dynamic cost by suppressing fast or aggressive motions, rather than by improving their closed-loop execution.

For example, suppose the policy predicts a fast upward motion in the $+z$ direction. If the controller cannot track this motion from the nominal command alone, directly penalizing the resulting tracking residual would encourage the policy to reduce the upward acceleration. This can be undesirable, since the acceleration may be essential for completing the task. Instead, if possible, we would like to preserve the intended reference if the controller could track it with the help of a bounded feed-forward command offset.
\begin{equation}
 J_{\mathrm{dyn}}^{\star} =\min_{o \in \mathcal{B}} J_{\mathrm{dyn}}
\big(
\phi({x + o}, q_0; \kappa),
x
\big)
+
\lambda_{\mathrm{reg}}\,\mathcal{R}(o).
\label{eq:jdyn_offset}
\end{equation}
Here, $x + o$ denotes the offset command sent to the controller, while the cost $J_{\mathrm{dyn}}$ is still evaluated against the original, unshifted reference $x$. The regularizer $\mathcal{R}(o)$ penalizes undesirable offsets, such as large or non-smooth corrections, and $\mathcal{B}$ bounds the admissible offset magnitude.

The resulting cost $J_{\mathrm{dyn}}^{\star}$ measures the tracking error that remains after the best bounded feed-forward correction has been applied. Thus, the guidance signal captures the part of the reference that cannot be absorbed by the controller, instead of rewarding the policy for simply slowing down the motion.

\subsubsection{Implementation Details}
\label{app:implementation_details}
\textbf{Kinematic feasibility model.}
For the grasping experiments, we approximate the distance-to-reachability cost \(J_{\mathrm{IK}}\) with a differentiable surrogate \(\hat{J}_{\mathrm{IK}}\), implemented as a small MLP trained on reachable and unreachable task-space targets for each robot embodiment. During optimization, \(\hat{J}_{\mathrm{IK}}\) is used in place of the ideal distance in Eq.~\eqref{eq:Jik}, making the feasibility cost cheap to evaluate and differentiable with respect to the generated grasp representation. Network architecture, training-data generation, and held-out accuracy are reported in Appendix~\ref{app:ik_surrogate}. \\
\\\textbf{Constrained NLP solver.} For the constrained formulation, we use IPOPT~\citep{ipopt} with a maximum of 45 iterations. The optimization variables are the initial denoising state \(x_K\) and the correction variables \(\{\delta_k\}_{k=1}^{K}\). The formulation is \begin{align} \label{eq:ipopt_opt} \begin{aligned} \min_{x_K,\,\{\delta_k\}_{k=1}^{K}} \quad & L_{\delta} + L_{\mathrm{IK}} \\ \mathrm{s.t.} \quad & x_{k-1} = \mu_\theta(x_k,k) + \sigma_k \delta_k, && k=K,\dots,1, \\ & J_{\mathrm{IK}}(x_0) \leq \varepsilon_{\mathrm{IK}}, \\ & -1 \leq x_K \leq 1, \\ & -1 \leq \delta_k \leq 1, && k=1,\dots,K , \end{aligned} \end{align} with \(\varepsilon_{\mathrm{IK}}=0.01\). The objective terms are \begin{align} L_{\delta} &= \lambda_{\delta} \sum_{k=1}^{K}\|\delta_k\|_2^2, \\ L_{\mathrm{IK}} &= \lambda_{\mathrm{IK}} \sum_{k=0}^{K} J_{\mathrm{IK}}(x_k), \end{align} where \(\lambda_{\delta}=1.0\) and \(\lambda_{\mathrm{IK}}=1.0\). When additional task constraints are present, further cost terms can be added to the objective. For example, we include an object-penetration penalty \(L_{\mathrm{obj}} = \lambda_{\mathrm{obj}} \sum_{k\in\mathcal{K}_{\mathrm{obj}}} P(x_k)\), with \(\lambda_{\mathrm{obj}}=1.0\) and \(\mathcal{K}_{\mathrm{obj}}=\{0,1\}\), where \(P(x_k)\) discourages hand--object interpenetration during the final denoising steps. \\

\newpage
\textbf{Differentiable NLS relaxation.} For the differentiable relaxation, we use Theseus with 15 Levenberg--Marquardt iterations. Instead of enforcing the reverse dynamics and terminal feasibility as hard constraints, we optimize the soft objective \begin{align} L_{\mathrm{NLS}} &= L_{\mathrm{rev}} + L_{\mathrm{rev},0} + L_{\delta} + L_{\mathrm{clamp}} + L_{\mathrm{IK,cost}} + L_{\mathrm{IK,term}} + L_{\mathrm{IK,path}}, \end{align} with \begin{align} L_{\mathrm{rev}} &= 10.0 \sum_{k=1}^{K} \left\| x_{k-1} - \mu_{\theta}(x_k,k) - \sigma_k\delta_k \right\|_2^2, \\ L_{\mathrm{rev},0} &= 50.0 \left\| x_0 - \mu_{\theta}(x_1,0) - \sigma_1\delta_1 \right\|_2^2, \\ L_{\delta} &= 1.0 \sum_{k=1}^{K} \|\delta_k\|_2^2, \\ L_{\mathrm{clamp}} &= 100.0 \sum_{k=1}^{K} \left\| \max(0,\delta_k-1) + \max(0,-1-\delta_k) \right\|_2^2, \\ L_{\mathrm{IK,cost}} &= 1.0 \sum_{k=1}^{K} J_{\mathrm{IK}}(x_k)^2, \\ L_{\mathrm{IK,term}} &= 10.0 \max\!\left(0,J_{\mathrm{IK}}(x_0)-0.005\right)^2, \\ L_{\mathrm{IK,path}} &= 10.0 \sum_{k=1}^{K} \max\!\left(0,J_{\mathrm{IK}}(x_k)-0.005\right)^2 . \end{align}
This relaxation preserves the same denoising structure as the constrained NLP but replaces hard feasibility with weighted penalties. \\

\textbf{Gradient-guidance baseline.} For the gradient-guidance baseline, we apply a per-step guidance update with scale \begin{align} s_k = g\,w(p_k) \frac{\beta_k}{\sqrt{1-\alpha_k}}, \end{align} where \(g=0.2\), \(p_k=1-k/K\), and \(w(p_k)=\cos^2(\pi p_k)\). The factor \(\beta_k/\sqrt{1-\alpha_k}\) accounts for the diffusion noise schedule. This baseline follows the same denoising schedule as the proposed method, but uses a scaled cost gradient instead of solving an optimization problem at each guided step. \\
Note that the gradient-guidance baseline uses a cosine schedule and the optimization-guided methods (IPOPT and Theseus) use constant schedulers. Both scheduler types were tested for all methods and the best performing one selected and reported. Similarly, various scales $g$ were tested, and $g=0.2$ yielded the best results.

\textbf{Solver settings.} Unless stated otherwise, the grasping experiments use IPOPT with 45 iterations and Theseus with 15 Levenberg--Marquardt iterations. For the image-conditioned dynamic manipulation experiments, we use the lightweight L-BFGS backend during the final denoising steps to keep online replanning tractable, which is detailed below.

\textbf{Visuomotor manipulation guidance.}
For the dynamic-manipulation experiments, we compare three ways of using the
controller-level executability cost \(J_{\mathrm{dyn}}\) from~\eqref{eq:Jdyn}.
All three methods use the same frozen diffusion model, replanning horizon, and
differentiable rollout model from Appendix~\ref{app:rollout}. The cost is
evaluated on the final denoised action chunk \(x_0\). The methods differ only
in how the correction enters the sampling process.

The first baseline is an \(x_t\)-nudge guidance method. At each denoising step,
we compute the guidance gradient \(g\) from \(J_{\mathrm{dyn}}\) and update
\begin{equation}
    x_t \leftarrow x_t + \lambda\,\bar\alpha_t\,g .
    \label{eq:xt_nudge}
\end{equation}
The standard DDIM update is then applied to the nudged sample. Since the
diffusion network receives the corrected \(x_t\), both the denoised estimate
\(\hat{x}_0(x_t)\) and the predicted noise
\(\epsilon_\theta(x_t,t,c)\) are affected by the correction. The factor
\(\bar\alpha_t\) concentrates the update near the final denoising steps. We use
\(\lambda=1\) and \(w_{\mathrm{dyn}}=5\).

The second baseline is \(\sigma\)-guidance, which leaves the network input
unchanged and injects the guidance gradient through the stochastic channel of
the DDIM update. After evaluating the network at the original \(x_t\), we compute
the deterministic DDIM mean \(\mu_t\) and set
\begin{equation}
    x_{t-1} = \mu_t + \sigma_t\,g ,
    \qquad
    \sigma_t =
    \eta\sqrt{
    \frac{1-\bar\alpha_{t-1}}{1-\bar\alpha_t}
    \left(1-\frac{\bar\alpha_t}{\bar\alpha_{t-1}}\right)} .
    \label{eq:sigma_guid}
\end{equation}
Its magnitude follows \(\sigma_t\), which is largest at noisy steps and vanishes
at the clean step. Therefore, this guidance is active only for \(\eta>0\); we use
\(\eta=1\).

In both guidance schemes, the guidance gradient \(g\) depends on the controller
offsets \(o\), which are themselves the solution of the inner optimization
problem in~\eqref{eq:jdyn_offset}. We therefore treat the correction as a
two-stage (bilevel) optimization: at each denoising iteration we first compute
the optimal offsets \(o\) with a few inner gradient steps, and then evaluate the
guidance gradient with these \(o\) held fixed.

The optimization-based variant uses direct-shooting L-BFGS. Instead of modifying
each denoising step online, it optimizes the full sampling process once per
replanning step. The decision variable contains the initial latent sample
\(x_T\), the injected denoising noises \(\{\varepsilon_i\}\), and the
feed-forward offset \(o\):
\[
    z = [\,x_T \mid \varepsilon \mid o\,] .
\]
We optimize \(z\) with \texttt{torch.optim.LBFGS} using learning rate \(0.1\),
at most \(10\) inner iterations, at most \(5\) closure evaluations, history size
\(10\), strong-Wolfe line search, and one outer step. Each closure re-runs the
full DDIM unroll with the frozen diffusion network, evaluates the differentiable
rollout on the terminal chunk \(x_0\), and back-propagates through both the
sampler and rollout. Since the direct-shooting parameterization explicitly
unrolls the diffusion dynamics, the generated trajectory remains consistent
with the sampler by construction.

\subsubsection{Base-pose difficulty stratification}
\label{app:basepose}
\begin{wrapfigure}{r}{0.7\textwidth}
    \centering
    \vspace{-5mm}
    \includegraphics[width=\linewidth]{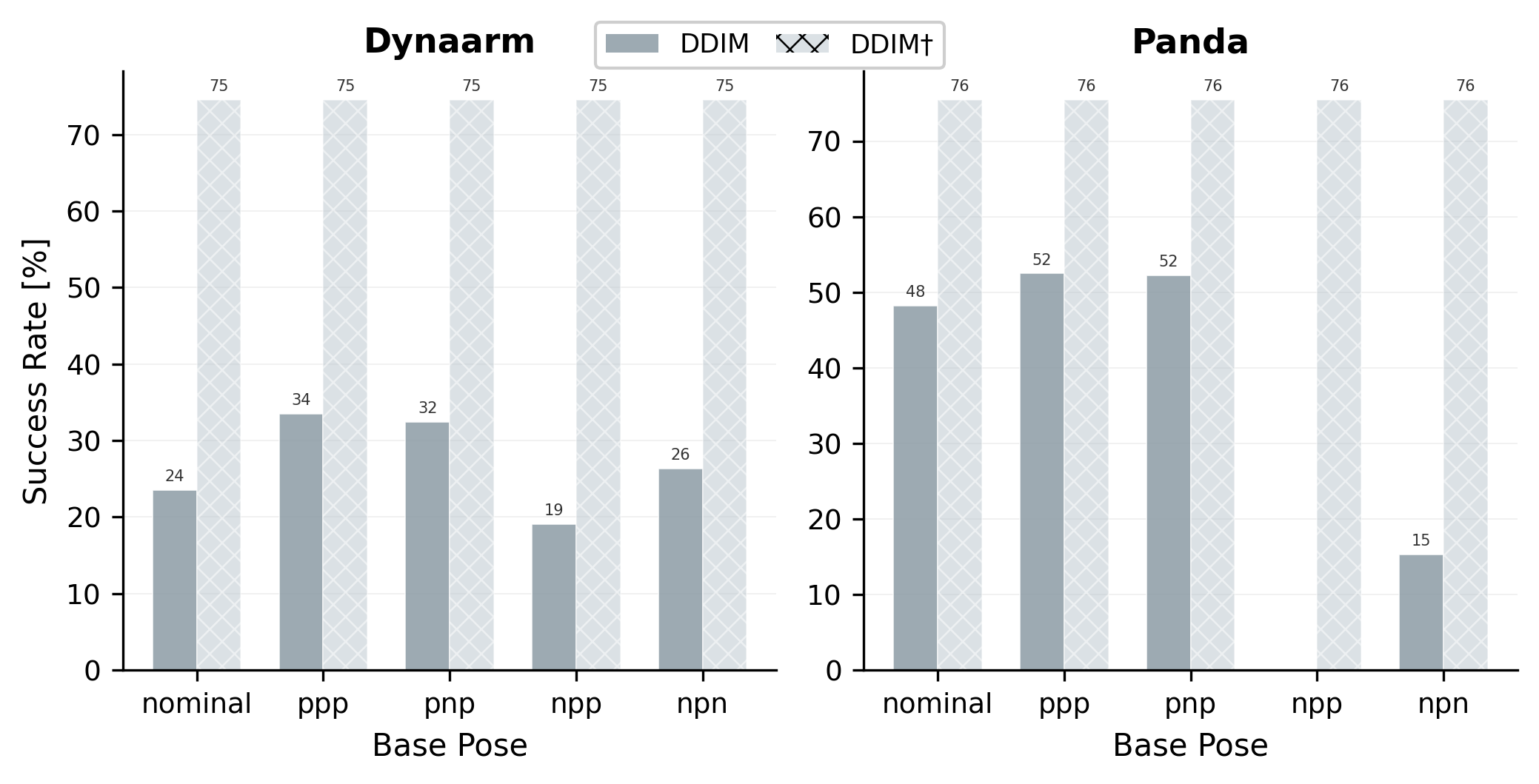}
    \caption{\textbf{Success rate by base-pose.} Per-method SR on the Dynaarm and Franka for each base-pose on a smaller subset of objects, with poses stratified by the success rate of the \textit{DDIM} approach.}
    \label{fig:ik_error_per_pose}
    \vspace{-2mm}
\end{wrapfigure}
We stratify the evaluation by base-pose difficulty: poses are classified as \textit{easy} or \textit{hard} based on the success rate of the \textit{DDIM} approach, which hard-projects the un-guided diffusion output \textit{DDIM}$\dagger$. Base poses with lower success rates are considered more difficult. Figure~\ref{fig:ik_error_per_pose} highlights the drop in success rate when projecting the diffusion output onto the inverse-kinematic feasible set, across both embodiments (Dynaarm and Franka).

\begin{figure*}[h]
    \centering
    \includegraphics[width=1.0\linewidth]{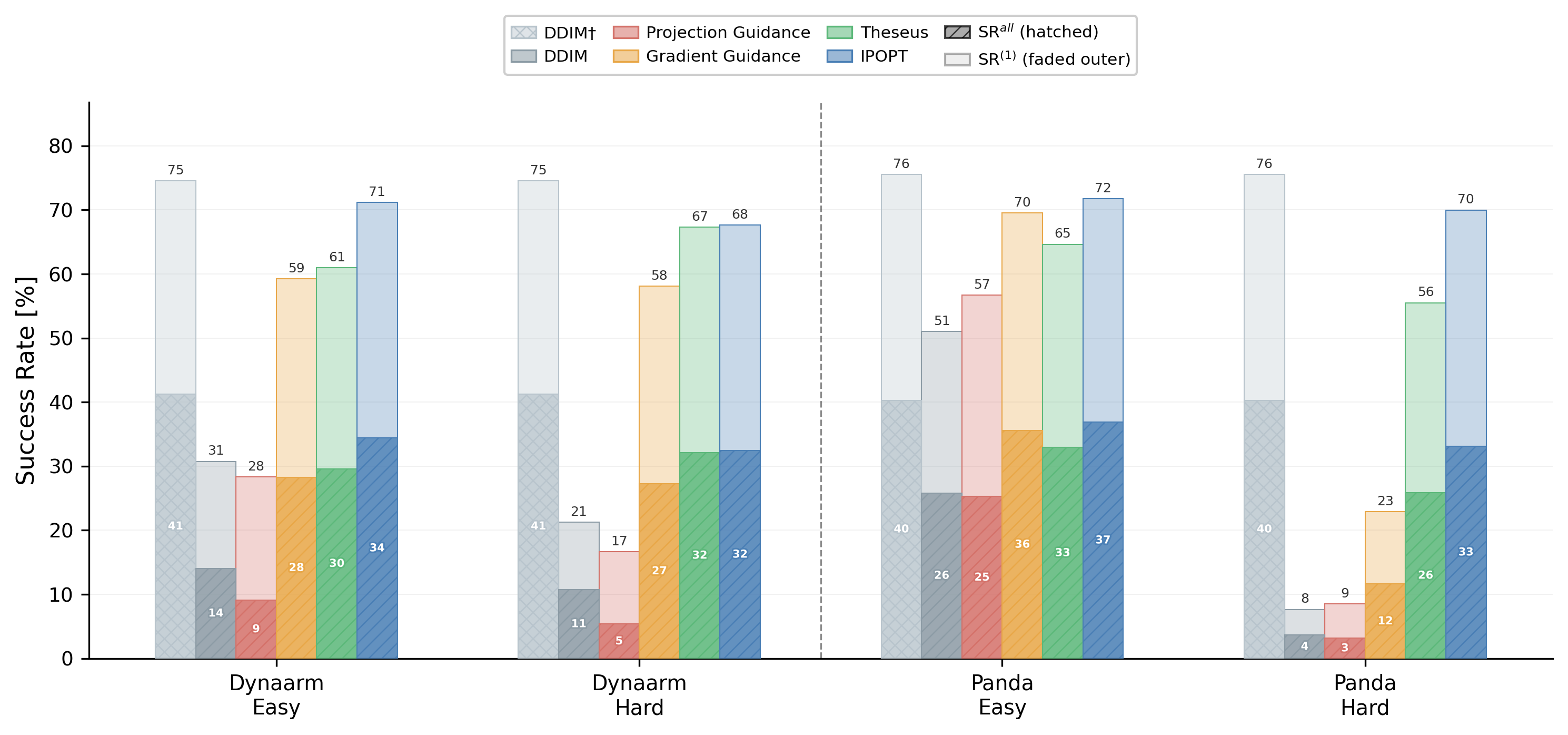}
    \caption{\textbf{Grasp success by base-pose difficulty.} Per-method success
    rates on the Dynaarm and Franka, split into \emph{easy} and \emph{hard}
    base-pose bins. Each bar overlays the per-pose success rate
    ($\mathrm{SR}^{(1)}$, faded outer) and the per-grasp success rate
    ($\mathrm{SR}^{\mathrm{all}}$, hatched inner). Our optimization-constrained
    variants (IPOPT, Theseus) dominate every (arm, difficulty) bin and degrade
    the least as poses harden.}
    \label{fig:easy_hard_poses}
\end{figure*}

Based on Figure~\ref{fig:ik_error_per_pose}, we consider poses \textit{nominal, npp} and \textit{npp, npn} as hard base pose configurations for the Dynaarm and Franka, respectively. The \textit{nominal} base pose is given as $[-0.65, 0, 0]$. The index \textit{p} refers to a shift in positive direction, \textit{n} in negative direction by $0.25$ from the \textit{nominal} pose. Pose \textit{npp} therefore refers to $[-0.65-0.25, 0.25, 0.25] = [-0.9, 0.25, 0.25]$.\\
Figure~\ref{fig:easy_hard_poses} reports per-pose ($\mathrm{SR}^{(1)}$, faded outer) and per-grasp ($\mathrm{SR}^{\mathrm{all}}$, hatched inner) success rates across both arms. $\text{DDIM}\dagger$ serves as an upper bound on achievable success rate given the trained diffusion prior. The trend of decreased success rates ($SR^{(1)}$) across all methods for the more challenging arm base poses is particularly evident for the Franka arm. Notably, both baselines (\textit{Gradient guidance} and \textit{Projection guidance}) suffer significant drops in success rate (from $70$ to $23$ and $57$ to $9$ respectively). The optimization-guided approaches (\textit{IPOPT} and \textit{Theseus}) are more robust to base pose changes with smaller drops in success rate (from $72$ to $70$ and $65$ to $56$ respectively). The Dynaarm shows similar trends, although less pronounced. \textit{Gradient guidance} and \textit{Projection guidance} drop from $59$ to $58$ and $28$ to $17$ while the optimization-guided \textit{IPOPT} drops from $71$ to $68$. Interestingly, \textit{Theseus} actually increases in success rate when changing from the \textit{easy} to \textit{hard} base pose configuration, going from $61$ to $67$. Additionally, while \textit{Gradient guidance} outperforms \textit{Theseus}, and almost approaches the success rate achieved by \textit{IPOPT}, on the Franka arm in the \textit{easy} base pose category, scores drop significantly on the \textit{hard} category, where \textit{Theseus} clearly outperforms \textit{Gradient guidance}.

\subsection{Evaluation Metrics}
\label{app:metrics}
We collect the definitions of all metrics used in the main text and in the
collision tables here.

\paragraph{Grasp success and feasibility.}
\begin{itemize}[leftmargin=1.5em] 
\item $\mathrm{SR}^{(1)}$ (\emph{one-axis grasp success}): a grasp is successful if it remains stable in simulation under at least one pull axis. To evaluate a grasp, we apply $3\,\mathrm{N}$ pulls along the $\pm x$, $\pm y$, and $\pm z$ axes of the object frame, grouped into three axis-wise phases. A grasp passes an axis if the object's center of mass remains within $5\,\mathrm{cm}$ of its initial position throughout that phase. \item $\mathrm{SR}^{\mathrm{all}}$ (\emph{all-axis grasp success}): a grasp is successful under the same evaluation if it passes all three axis-wise pull phases. This is the stricter counterpart of $\mathrm{SR}^{(1)}$, requiring the same grasp to remain stable under pulls along all object-frame axes.
\item $\mathrm{SR}^{\mathrm{IK}}$ (\emph{kinematic feasibility}): the fraction of predictions that are reachable by the target arm. For the snap-to-feasible DDIM baseline, this metric is computed \emph{before} projection (Footnote~\ref{fn:srik_projection}). \item $\mathrm{CR}$ (\emph{collision rate}, collision tables only): the fraction of predicted grasps whose hand geometry penetrates an obstacle by more than $3\,\mathrm{mm}$. We compute hand surface points from the predicted grasp pose and query their signed distances to the obstacle SDFs. A grasp is marked as colliding if the maximum penetration over all hand points and obstacles exceeds the tolerance. This is a purely geometric check and does not involve physics simulation. \item $\mathrm{SR}^{\mathrm{Tot}}$ (\emph{total success}, collision tables only): the fraction of predictions that are both grasp-successful and collision-free. We compute this by zeroing out the per-grasp success indicator for any grasp in collision, and then averaging over all grasps. \end{itemize}

\paragraph{Grasp quality.} 
\begin{itemize}[leftmargin=1.5em]
    \item $E_{\mathrm{fc}}$ (\emph{force-closure violation}): the GraspQP span-metric residual~\citep{graspqp2025}, capturing how close the grasp is to force closure (lower is better). In practice, we use the method of \citep{zurbruegg2026dexevolve} to resample contact points and rely on the four-sided friction pyramid using $\mu=0.2$.
    \item $Q_1$ (\emph{grasp wrench quality}, higher is better): the Ferrari-Canny $\varepsilon$-metric, given by the radius of the largest origin-centered ball inside the convex hull of the grasp wrench set. 
\end{itemize}

\subsection{Collision-environment results}
\label{app:collision}
For the collision-aware experiments, we use the same diffusion architecture as in
the main grasping experiments, but train it on the full Dataset from~\citep{graspqp2025} to
improve robustness under the tighter scene constraints and to support
sim-to-real transfer. In addition, the initial wrist poses $x_K$ are sampled above the
table plane to warm-start diffusion, providing a better initialization from which gradient-based methods can more easily find solutions.

Table~\ref{tab:collision} evaluates collision-aware grasp generation across
four constrained environments: floor, walls, clutter, and tunnels. The
\textit{walls} setting places a wall perpendicular to the arm base, whereas the
\textit{tunnel} setting places walls on both sides of the arm, forming a
funnel-like constraint. All methods are evaluated on $4$ objects with $10$
grasps per object in each environment. Since these scenes substantially restrict
the set of valid grasps, the experiment stresses the ability of each method to
balance grasp quality, kinematic reachability, and collision avoidance.

We report the single-grasp success rate $\mathrm{SR}^{(1)}$, inverse-kinematics
success rate $\mathrm{SR}^{\mathrm{IK}}$, collision rate $\mathrm{CR}$, and total
success rate $\mathrm{SR}^{\mathrm{Tot}}$, where total success denotes grasps
that both succeed and remain collision-free. All guided methods use a proximity
cost based on the signed distance field (SDF),
\begin{align}
    J_{\mathrm{coll}}(x) = \max(0, s(x) + d_{\mathrm{safe}}),
\end{align}
with safety margin $d_{\mathrm{safe}}=5\,\mathrm{cm}$. This formulation provides a
non-zero optimization signal before actual obstacle penetration occurs. For
IPOPT, we include $J_{\mathrm{coll}}$ as a cost term rather than a hard
constraint to avoid overly restrictive local optimization landscapes.

\begin{table*}[h]
\centering
\caption{\textbf{Grasp Predictions with Collision Avoidance:} Comparison
of methods across evaluation metrics (averaged over $4$ objects and $10$
grasps per object) for $4$ different environments, with the walls oriented
perpendicular to the arm base.}
\label{tab:collision}
\resizebox{\textwidth}{!}{%
\setlength{\tabcolsep}{4pt}
\begin{tabular}{l|l||c|rrr||c|rrr||c|rrr||c|rrr}
\toprule
\multirow{4}{*}{\textbf{Arm}} 
& \multirow{4}{*}{\textbf{Method}} 
& \multicolumn{16}{c}{\textbf{Environment}}\\
\cmidrule(lr){3-18}
& & \multicolumn{4}{c|}{\textbf{Floor}}
& \multicolumn{4}{c|}{\textbf{Walls}} 
& \multicolumn{4}{c|}{\textbf{Clutter}}
& \multicolumn{4}{c}{\textbf{Tunnels}} \\
\cmidrule(lr){3-6} \cmidrule(lr){7-10} \cmidrule(lr){11-14} \cmidrule(lr){15-18}
& 
& $\text{SR}^{\text{Tot}}${\scriptsize$\uparrow$}
& $\text{SR}^{(1)}$ {\scriptsize$\uparrow$} 
& $\text{SR}^{\text{IK}}$ {\scriptsize$\uparrow$} 
& $\text{CR}$ {\scriptsize$\downarrow$}
& $\text{SR}^{\text{Tot}}${\scriptsize$\uparrow$}
& $\text{SR}^{(1)}$ {\scriptsize$\uparrow$} 
& $\text{SR}^{\text{IK}}$ {\scriptsize$\uparrow$} 
& $\text{CR}$ {\scriptsize$\downarrow$}
& $\text{SR}^{\text{Tot}}${\scriptsize$\uparrow$}
& $\text{SR}^{(1)}$ {\scriptsize$\uparrow$} 
& $\text{SR}^{\text{IK}}$ {\scriptsize$\uparrow$} 
& $\text{CR}$ {\scriptsize$\downarrow$}
& $\text{SR}^{\text{Tot}}${\scriptsize$\uparrow$}
& $\text{SR}^{(1)}$ {\scriptsize$\uparrow$} 
& $\text{SR}^{\text{IK}}$ {\scriptsize$\uparrow$} 
& $\text{CR}$ {\scriptsize$\downarrow$}\\
\midrule
\multirow{5}{*}{Dynaarm}
& DDIM~\citep{zurbruegg2026dexevolve}              
& $5.0$ & $16.25$ & $32.5$ & $60.0$ 
& $2.5$ & $16.25$ & $32.5$ & $62.5$
& $4.37$ & $19.99$ & $32.5$ & $67.5$ 
& $0.0$ & $16.25$ & $32.5$ & $80.0$ \\
\rowcolor{gray!10} & Gradient Guidance
& $33.12$ & $33.12$ & $100$ & $0.0$ 
& $17.5$ & $19.99$ & $100$ & $25.0$ 
& \underline{$28.12$} & $28.12$ & $100$ & $0.0$
& $\mathbf{30.63}$ & $30.62$ & $100$ & $0.0$  \\
& Projection Guidance
& $19.38$ & $22.49$ & $100$ & $20.0$ 
& $0.63$ & $5.62$ & $100$ & $20.0$ 
& $0.0$ & $0.0$ & $75.0$ & $0.0$  
& $14.37$ & $19.37$ & $100$ & $22.5$ \\
\cmidrule(l){2-18}
\rowcolor{gray!10} 
& Theseus
& \underline{$36.88$} & $36.87$ & $77.5$ & $0.0$
& $\mathbf{35.63}$ & $44.36$ & $77.5$ & $17.5$
& $23.13$ & $28.37$ & $82.5$ & $37.5$
& \underline{$24.37$} & $46.24$ & $80.0$ & $32.5$ \\
& IPOPT
& $\mathbf{46.88}$ & $51.24$ & $75.0$ & $10.0$  
& \underline{$22.5$} & $43.74$ & $75.0$ & $32.5$ 
& $\mathbf{31.25}$ & $39.99$ & $65.0$ & $20.0$ 
& $23.75$ & $26.24$ & $52.5$ & $32.5$ \\
\midrule
\midrule
\multirow{5}{*}{Franka}
& DDIM~\citep{zurbruegg2026dexevolve}
& $11.87$ & $68.73$ & $100$ & $82.5$ 
& $10.0$ & $68.73$ & $100$ & $85.0$ 
& $6.25$ & $60.61$ & $100$ & $90.0$
& $7.5$ & $68.73$ & $100$ & $87.5$ \\
\rowcolor{gray!10} 
& Gradient Guidance
& $22.5$ & $22.49$ & $100$ & $0.0$
& $20.62$ & $20.62$ & $100$ & $30.0$ 
& $23.13$ & $23.74$ & $100$ & $17.5$ 
& $18.12$ & $18.12$ & $100$ & $5.0$ \\
& Projection Guidance
& $58.13$ & $60.61$ & $100$ & $2.5$ 
& $43.13$ & $45.61$ & $100$ & $2.5$
& $0.0$ & $0.0$ & $57.5$ & $0.0$ 
& \underline{$45.0$} & $47.49$ & $100$ & $10.0$ \\
\cmidrule(l){2-18}
\rowcolor{gray!10} 
& Theseus
& \underline{$58.75$} & $58.74$ & $100$ & $0.0$
& \underline{$50.62$} & $55.61$ & $100$ & $17.5$
& \underline{$26.87$} & $38.12$ & $97.5$ & $27.5$
& $38.75$ & $59.36$ & $100$ & $30.0$ \\
& IPOPT
& $\mathbf{65.62}$ & $69.98$ & $100$ & $12.5$ 
& $\mathbf{60.62}$ & $76.86$ & $100$ & $20.0$
& $\mathbf{55.63}$ & $70.61$ & $100$ & $27.5$ 
& $\mathbf{61.88}$ & $68.11$ & $100$ & $17.5$ \\
\bottomrule
\end{tabular}
}
\end{table*}
Across both embodiments, the optimization-based methods obtain the best total
success in most environments. On the Franka, IPOPT achieves the highest
$\mathrm{SR}^{\mathrm{Tot}}$ in all four environments, reaching $65.62\%$ on
\textit{Floor}, $60.62\%$ on \textit{Walls}, $55.63\%$ on \textit{Clutter}, and
$61.88\%$ on \textit{Tunnels}. On the Dynaarm, IPOPT and Theseus also improve
over the baselines in several constrained settings, with IPOPT performing best
on \textit{Floor} and Theseus on \textit{Walls}. The main exception is the
\textit{Tunnels} environment, where Gradient Guidance achieves the highest
$\mathrm{SR}^{\mathrm{Tot}}$ for the Dynaarm. In this case, the local gradient
signal appears to provide a favorable correction direction, reducing collisions
without substantially degrading grasp quality.

However, this behavior is not consistent across environments or embodiments.
The same gradient-based update often satisfies one objective at the expense of
another. For example, Gradient Guidance achieves a $0.0\%$ collision rate in
several Dynaarm environments, but this frequently comes with reduced grasp
success. Similarly, Projection Guidance reaches a $0.0\%$ collision rate in the
\textit{Clutter} environment for both arms, but also collapses to $0.0\%$
grasp success, which corresponds in simulation to grasps being moved into free
space rather than onto the object. These cases indicate that local gradient or
projection updates can work well when the correction direction is aligned with
the learned grasp prior, but can also move samples away from successful grasp
modes when the objectives conflict. In contrast, the optimization-based
formulations provide a more robust mechanism for trading off grasp quality,
reachability, and collision avoidance, leading to higher total success in most
of the evaluated settings.

\subsubsection{Real-World Deployment}
\label{app:real_world}

\begin{figure*}[t]
    \centering
    \includegraphics[width=1.0\linewidth]{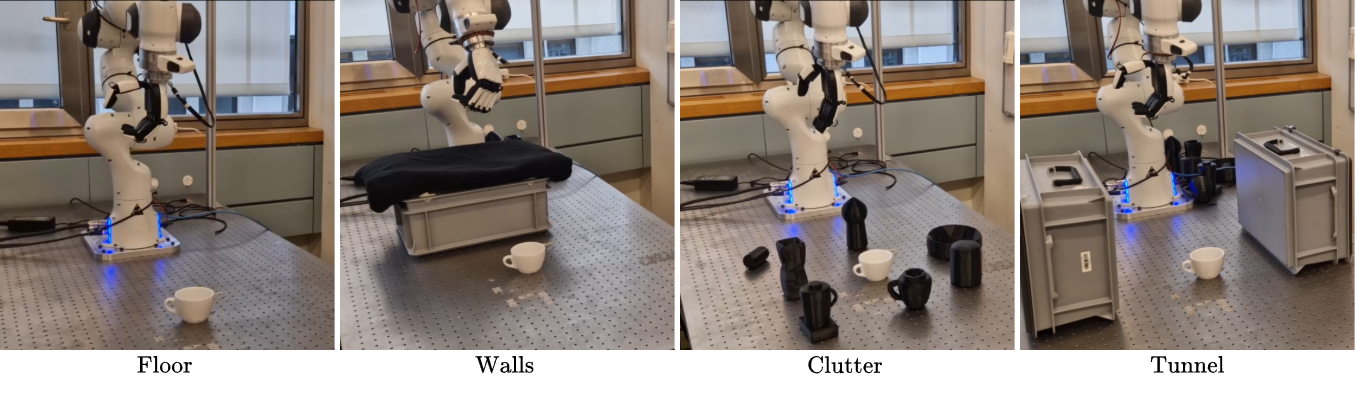}
\caption{\textbf{Hardware Setup.}
Illustration of the four environments considered during hardware deployment. A video of all hardware experiments is provided in the supplementary material.}
    \label{fig:hardware_setups}
\end{figure*}

The dexterous manipulation task was additionally deployed on hardware using a Franka Panda arm equipped with the XHand. We evaluated both the gradient-based guidance baseline and our IPOPT-based method across all four environment settings, \textit{Floor}, \textit{Walls}, \textit{Clutter}, and \textit{Tunnels}, shown in Fig.~\ref{fig:hardware_setups}. While the simulation study in Table~\ref{tab:collision} considers four objects, we restrict the hardware evaluation to two representative objects, \textit{Mug} and \textit{Spices}. Both methods are deployed zero-shot, without retraining or fine-tuning of the diffusion model. \underline{Additional hardware rollouts are shown in the supplementary video.}

For the hardware evaluation, we generated multiple grasp candidates per method, object, and environment. In the \textit{Floor} environment, we generated three grasps for each method-object pair. For the more constrained
\textit{Walls}, \textit{Clutter}, and \textit{Tunnels} environments, we generated five grasps per method-object pair to account for the smaller set of valid solutions. We then deployed all candidates that were executable by the hardware pipeline.

We note that the arm approach motion was planned using cuRobo, which performs collision avoidance during execution. This component is part of the deployment pipeline, but not part of our proposed method or evaluation claim: our method only optimizes the final grasp configuration with respect to grasp success,
kinematic reachability, and environment collisions. Consequently, some generated grasps that satisfied the final-state criteria could not be deployed because no valid collision-free approach motion was found by the controller. We therefore interpret the hardware experiments as initial deployment evidence rather than a complete end-to-end benchmark of motion planning and grasp execution.
\newpage

\begin{wrapfigure}{r}{0.5\textwidth}
    \centering
    \includegraphics[width=\linewidth]{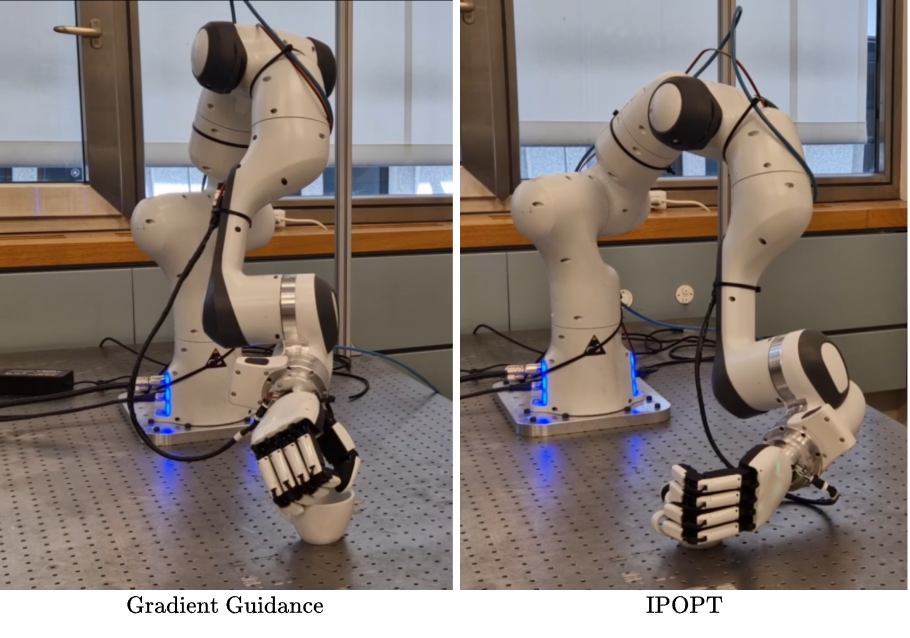}
    \caption{\textbf{Out-of-distribution failure case.}
    Example of gradient guidance and IPOPT in the \textit{Floor} environment.
    Gradient guidance respects the collision-avoidance objective with respect
    to the floor, but fails to grasp the object.}
    \label{fig:hardware_floor_failure}
\end{wrapfigure}
Within this protocol, the hardware experiments support the qualitative trend observed in Table~\ref{tab:collision}. In particular, gradient guidance can produce collision-free final grasps while moving the hand away from useful
object-centered grasp modes, leading to missed or unstable grasps even in the comparatively simple \textit{Floor} environment, as illustrated in Fig.~\ref{fig:hardware_floor_failure}. In contrast, the IPOPT-based method more reliably preserved grasp structure in the deployed trials while accounting for the environmental constraints, as also shown in Fig.~\ref{fig:hardware_floor_failure}. These results provide initial evidence that the same optimization-guided corrections used in simulation can transfer zero-shot to physical deployment.\\
Further hardware experiments for all four environments can be found in the supplementary video.